\documentclass[review]{elsarticle}

\usepackage{bm}
\usepackage{lineno,hyperref}
\usepackage{mathtools}
\usepackage{amsthm,nccmath}

\usepackage{graphicx}
\usepackage{multirow}

\usepackage{algorithm}
\usepackage{algpseudocode}
\usepackage{algpascal}
\usepackage{textcomp}
\usepackage{tikz}

\usepackage{smartdiagram}
\usetikzlibrary{arrows,positioning}
\usepackage{makecell} % for more vertical space in cells
\setcellgapes{5pt}

\usepackage{amsmath,amsfonts,amssymb}

\DeclareMathOperator*{\argmin}{arg\,min}

\smartdiagramset{border color=none,
	set color list={blue!50!cyan,green!60!lime,orange!50!red,red!80!black},
	back arrow disabled=true, module minimum width = 4}

\smartdiagramset{module minimum height=1cm,% initial: 1cm
	module minimum width=0.8cm,% initial: 2cm
	module x sep=2}% initial 2.75

\usepackage{lineno,hyperref}
\modulolinenumbers[5]

\usepackage{xcolor}
\begin{document}
	\fontsize{10}{9}\selectfont
	%\sloppy
	
	\begin{frontmatter}
		
		\title{Time series clustering based on prediction accuracy of global forecasting models}

		\author[mymainaddress]{\'Angel L\'opez-Oriona\corref{mycorrespondingauthor} (ORCID 0000-0003-1456-7342)}
		\ead{oriona38@hotmail.com, a.oriona@udc.es}
		
		\author[dursoaddress]{Pablo Montero-Manso (ORCID 0000-0003-3816-0985)}
		\ead{pablo.monteromanso@sydney.edu.au}
		
		\author[mymainaddress]{Jos\'e A. Vilar (ORCID 0000-0001-5494-171X)}
		\cortext[mycorrespondingauthor]{Corresponding author}
		\ead{jose.vilarf@udc.es}
		
		\address[mymainaddress]{Research Group MODES, Research Center for Information and Communication Technologies (CITIC), University of A Coru\~na, 15071 A Coru\~na, Spain.}
		\address[dursoaddress]{The University of Sydney Business School, Australia.}
		\begin{abstract}
  %{**NOTES: Remove the K-means reference, add model selection, modify claims of optimality, add key ideas**}\\
			In this paper, a novel method to perform model-based clustering of time series is proposed. The procedure relies on two iterative steps: (i) $K$ global forecasting models are fitted via pooling by considering the series pertaining to each cluster and (ii) each series is assigned to the group associated with the model producing the best forecasts according to a particular criterion. Unlike most techniques proposed in the literature, the method considers the predictive accuracy as the main element for constructing the clustering partition, which contains groups jointly minimizing the overall forecasting error. Thus, the approach leads to a new clustering paradigm where the quality of the clustering solution is measured in terms of its predictive capability. In addition, the procedure gives rise to an effective mechanism for selecting the number of clusters in a time series database and can be used in combination with any class of regression model. An extensive simulation study shows that our method outperforms several alternative techniques concerning both clustering effectiveness and predictive accuracy. The approach is also applied to perform clustering in several datasets used as standard benchmarks in the time series literature, obtaining great results.
		\end{abstract}
		
		
	\end{frontmatter}

	\section{Introduction}\label{sectionintroduction}
	
	Time series clustering (TSC) is a fundamental problem in machine learning with applications in many fields, including biology, economics, computer science or psychology, among others. The task consists of splitting a large collection of unlabelled time series realizations into homogeneous groups so that similar series are located together in the same group and dissimilar series are placed in different clusters. As result, each group can be characterized by a specific temporal pattern, which allows to address key issues as discovering hidden dynamic structures, identifying anomalies or forecasting future behaviours. Comprehensive overviews on the topic are provided in \cite{Liao:2005, fu2011review, Rani_Sikka:2012, aghabozorgi2015time, maharaj2019time}.  
	
	A crucial point in cluster analysis is to establish the dissimilarity notion since it determines the nature of the resulting clustering partition. Several distance measures have been proposed in the literature, each one of them associated with a different objective. If the goal is to discriminate between geometric profiles of the time series, then a shape-based dissimilarity is suitable. For instance, the well-known dynamic time warping (DTW) distance has been used in several works to perform TSC \cite{oates1999clustering, izakian2015fuzzy, wang2018time, d2021trimmed}. On the contrary, a structure-based dissimilarity is desirable if the target is to compare underlying dependence models. Examples of this type of distances are metrics comparing the autocorrelations \cite{d2009autocorrelation}, the quantile autocovariances \cite{lafuente2016clustering}, the quantile cross-spectral densities \cite{lopez2021quantile,lopez2022quantilea, lopez2022quantileb, lopez2021spatial}, the wavelet representations \cite{maharaj2010wavelet} or the wavelet coefficients \cite{d2020cepstral} of two time series. Additional types of dissimilarities are based on dimensionality reduction techniques \cite{yang2004pca, singhal2005clustering} or levels of shared information \cite{keogh2004towards, brandmaier2011permutation}. 
	
	Among TSC, the so-called model-based clustering is a popular approach which gave rise to several works. These techniques rely on two main elements: (i) the assumption of the existence of a fixed number of models characterizing the different groups in the time series dataset and (ii) a practical procedure to partition the series in a suitable way according to the underlying models. In an early work, \cite{piccolo1990distance} proposed to perform TSC by employing a distance measure which is based on the ARIMA representation of the time series. A similar method was introduced by \cite{d2013clustering} for clustering financial time series. Specifically, the technique assumes the existence of different GARCH models and uses a metric based on estimated GARCH parameters. The assumption of underlying GARCH models is also used by \cite{d2016garch} to construct different robust methods based on unconditional volatility and time-varying volatility of the GARCH representation of the time series. A novel mixture model for clustering series which are subject to regime changes was proposed by \cite{same2011model}. Particularly, the approach consists of modeling each cluster by a regression model in which the polynomial coefficients vary according to a discrete hidden process. It is worth remarking that this method belongs to a paradigm called clusterwise regression, which is based on considering that the elements within each cluster are generated according to a specific linear regression scheme \cite{hennig1999models}. In the multivariate setting, \cite{maharaj1999comparison} introduced a clustering method based on the $p$-value of a test of hypothesis assuming linear models, and \cite{frohwirth2008model} proposed to pool multiple time series into several groups using finite-mixture models, documenting the efficiency gains in estimation and forecasting realized relative to the overall pooling of the time series. In the categorical context, \cite{fruhwirth2010model} constructed two clustering approaches based on time-homogeneous first-order Markov chains. 
	
	Note that, although the previous techniques for model-based clustering of time series attempt to identify the underlying models existing in a given dataset, they ignore the performance of these models in terms of predictive accuracy. In this context, the aim of this manuscript is to propose a model-based clustering approach producing a clustering solution with a high predictive accuracy. Our idea is motivated by the fact that, given two different model-based clustering solutions, the one generating the best predictions is preferred. In short, our approach is able to detect the underlying models while trying to optimize the predictive accuracy. To that aim, we assess the dissimilarity between a time series and a given model as the average prediction error produced when iteratively obtaining the point forecasts of the time series with respect to the corresponding model. It is worth highlighting that, although there are a few TSC methods based on forecast densities \cite{alonso2006time, vilar2010non}, to the best of our knowledge, nobody has employed the concept of similarity mentioned above to perform clustering in time series databases. Specifically, our clustering approach makes use of the so-called global models (see Section \ref{sectionbackgroundglobalmodels}) to minimize the average prediction error. In fact, the use of global models circumvents some limitations that one often faces when fitting a different model to each time series in the set, i.e., when considering the so-called local approach. For instance, the predictive accuracy of these independent models is often poor when dealing with short time series, but this is not the case for global models. 
	
   Based on previous comments, we propose a novel clustering method which is based on traditional iterative clustering algorithms. The technique relies on the following iterative process: (i) $K$ global models (prototypes) are fitted by taking into account the series pertaining to each cluster independently and (ii) each time series is assigned to the group associated with the prototype producing the lowest forecasting error according to a specific metric.
	
	It is worth emphasizing that, by construction, the proposed algorithm produces a partition which is optimal in terms of overall prediction accuracy. In fact, the objective function of the method can be seen as a sum of forecasting errors (see Remark 1 in Section \ref{sectionclusteringalgorithm}), which is expected to decrease with each iteration of the two-step procedure described above. Therefore, the clustering algorithm is specifically designed to allocate the different time series in such a way that the corresponding global models represent in the best possible manner the existing prediction patterns. There are only a few works in the literature combining clustering and global methods in a single technique. For instance, \cite{bandara2020forecasting} proposed an approach particularly devised to improve the predictive accuracy of global models. First, the set of series is partitioned into different groups by using a specific clustering method. Then, global models are fitted by considering the series within each cluster. Although successful, the method of \cite{bandara2020forecasting} splits the set of series by using a feature-based TSC clustering approach and, therefore, there is not guarantee that the resulting partition is optimal in terms of total prediction accuracy. Note that our approach circumvents this limitation by adapting the objective function to the specific purpose of forecasting error reduction. It is important to highlight that, although an improvement in the overall predictive effectiveness is usually achieved through the proposed method, the main output of the procedure is the resulting clustering partition, which produces a meaningful decomposition of the set of time series in terms of forecasting structures and can be very useful as a exploratory tool. 
	
	Some simulation experiments are carried out in the paper to assess the performance of the proposed algorithm in terms of both clustering effectiveness and . In all cases, synthetic partitions where the groups are characterized by different generating processes are considered. The approach is compared with several alternative methods, as one procedure based on local models or the technique of \cite{bandara2020forecasting}. Several elements are analysed, including the type of global models, the way in which the series are assigned to the clusters or the numerical behaviour of the algorithm. The method is also applied to perform clustering in some well-known datasets which are used as classical benchmarks in the time series literature. Overall, the algorithm exhibits a great behaviour when dealing with both synthetic and real data. 
	
	An overview of the contributions provided in this manuscript is given below:
	
	\begin{itemize}
	\item The proposed approach exhibits a great ability to detect the underlying structures in several simulation experiments including different types of generating processes. In particular, we consider linear models with short and long memory and specific types of nonlinear processes. In most cases, the method outperforms the local approach and other alternatives in terms of clustering effectiveness, thus taking advantage of the underlying ability of global models to identify the different prediction patterns. 
	\item Our method provides an effective and natural way of automatically determining the number of clusters, which is an important topic in the TSC literature. Specifically, as the objective function of the algorithm can be seen as a sum of prediction errors, one can select the number of groups by choosing the value which minimizes a proper generalization of this objective function. Several experiments demonstrate that the true number of clusters is frequently selected by means of this procedure. 
	\item Generally, the proposed technique improves the overall predictive performance of a collection of time series in comparison with both the local approach and the consideration of one single global model. Specifically, each one of the groups has an associated model exploiting all the information about the corresponding forecasting structure, which makes that model the best choice to predict future values of the time series in the group. This improvement in predictive accuracy is corroborated by means of some simulations and experiments with well-known real time series datasets which are often used for forecasting purposes.  
	\end{itemize}

    It is worth highlighting that the proposed approach has also some limitations. First, the class of global models can have a great impact on the identification of the true clustering structure. In fact, for a proper identification, it is necessary that the complexity of the global models matches the underlying forecasting structures. In this regard, more complexity leads generally to fewer clusters, while the opposite happens with less complexity. Second, when the generating processes are not too complex (e.g., linear models with short memory), then the local approach reaches similar results than our method when moderate values of the series length are considered, since such lengths are enough for the coefficients of the local models to be estimated with high accuracy. Third, the proposed algorithm decreases its performance when some amount of uncertainty (noise) exists in the underlying structures, that is, when the time series dataset does not contain totally well-defined clusters. Fourth, as the proposed iterative method considers the future parts of the series to calculate the distance between each element and each global model, some numerical issues arise in the behaviour of the objective function. However, these negative effects can be easily neutralized by means of a simple heuristic rule (see Remark 3 in Section \ref{sectionclusteringalgorithm}). 
	
	The remainder of this paper is organized as follows. Section \ref{sectionbackgroundglobalmodels} gives a brief background on global forecasting models, while Section \ref{sectionclusteringalgorithm} describes the clustering algorithm based on prediction accuracy of these models, which is motivated through an interesting example in Section \ref{sectionmotivatingexample}. The approach is analysed in Section \ref{sectionsimulationstudy} by means of a simulation study where different scenarios are taken into account. In Section \ref{sectionapplication}, we apply the proposed method to real datasets of time series belonging to different fields. Section \ref{sectionconclusions} contains some concluding remarks and future work. 
	
	\section{Background on global models}\label{sectionbackgroundglobalmodels}
	
	Global models are learning algorithms that fit the same forecasting function to all the time series in a set, in contrast to local models, which adjust a different function to each time series in the database \cite{montero2021principles}. Rigorously, let $\mathbb{X}$ be the collection of all sets of univariate time series of finite size, i.e.,
	
	\begin{equation}
		\begin{split}
			\mathbb{X}=\bigg\{\mathcal{X}: \mathcal{X}=\Big\{\bm X_t^{(1)}, \ldots, \bm X_t^{(r)}\Big\}, \text{with} \, \, r \in \mathbb{N} \, \, \text{and} \, \, \bm X_t^{(i)} \in \mathbb{R}^{T}, i=1,\ldots, r \bigg\},
		\end{split}
	\end{equation}
	
	\noindent where we assumed without loss of generality that all series have the same length $T$ (they are vectors in the space $\mathbb{R}^T$). Usually, we are interested in the future part of each series up to $h$ time steps, which can be seen as a vector of $\mathbb{R}^h$. To compute the corresponding predictions, we employ a forecasting function $f$, which is a function from the observed time series to the future part, i.e., $f: \mathbb{R}^T \longrightarrow \mathbb{R}^h$, often defined in an iterative way when $h>1$. A global method, $\mathcal{A}_G$, is a learning algorithm taking the form
	
	\begin{equation}
		\mathcal{A}_G: \mathbb{X} \longrightarrow \mathbb{F}_T^h,
	\end{equation}
	
	\noindent where $\mathbb{F}_T^h$ is the set of all functions with domain $\mathbb{R}^T$ and range $\mathbb{R}^h$. Note that, for each set of series $\mathcal{X} \in \mathbb{X}$, $\mathcal{A}_G(\mathcal{X})$ defines a forecasting function created by using all the series in $\mathcal{X}$. In this paper we consider global models constructed in the following way \cite{montero2021principles}: (i) each series in $\mathcal{X}$ is lag-embedded into a matrix at a given autoregressive (AR) order, $l$, fixed beforehand, (ii) these matrices are stacked together to form one big matrix, achieving data pooling and (iii) a classical regression model (e.g., linear regression, random forest etc) is fitted to the resulting matrix.
	
	Global models have been shown to outperform local models in terms of predictive accuracy in several datasets \cite{montero2021principles}. In other words, when a single model is fitted to all the time series in the database, and used to obtain the corresponding predictions, a lower overall forecasting error is produced than in the case where each time series is predicted by considering a different local model. Moreover, global models do not need any assumption about similarity of the time series in the collection, and usually require far fewer parameters than the simplest of local methods. 
	
	Although the global approach produces outstanding results, it has one important drawback: it ignores the possible existence of homogeneous groups of series in terms of prediction patterns. For instance, a database could contain two groups of series in such a way that the series within each group are helpful to each other for obtaining accurate predictions (e.g., think of several countries whose behaviour concerning monthly economic growth is very similar), but totally useless for the series in the remaining group. In the previous situation, it would be desirable to fit a global method for each distinct set of time series. Then the predictions would be computed for a given series by using its associated global model. This is the main idea beyond our clustering method based on prediction accuracy of global models, which is introduced in the next section. 
	
	\section{A clustering algorithm based on prediction accuracy of global forecasting models}\label{sectionclusteringalgorithm}
	
	Consider a set of $n$ time series, $\mathcal{S} = \left\{ \bm{X}_t^{(1)}, \ldots, \bm{X}_t^{(n)} \right\}$, where each $\bm{X}_t^{(i)}=\big(X_1^{(i)}, \ldots, X_{L_i}^{(i)}\big)$ is a series of length $L_i$, $i=1,\ldots,n$. We assume that each series $\bm{X}_t^{(i)}$ contains training and validation periods of lengths $r(i)$ and $s(i)$, denoted by $\bm{\mathcal{T}}^{(i)}=(t^i_1, \ldots, t^i_{r(i)})$ and $\bm{\mathcal{V}}^{(i)}=(v_1^i, \ldots,v_{s(i)}^i)$, respectively, such that:
	
	\begin{itemize}
		\item Both $\bm{\mathcal{T}}^{(i)}$ and $\bm{\mathcal{V}}^{(i)}$ are formed by consecutive observations and $t_1^i$ has a position equal to or less than the position of $v_1^i$, considering both $t_1^i$ and $v_1^i$ as elements of the vector $\bm{X}_t^{(i)}$.
		\item Both periods are included in the original series.
		\item Both periods form a cover of the original series.
	\end{itemize}
	
	The sets $\mathcal{T}=\{\bm{\mathcal{T}}^{(1)}, \ldots, \bm{\mathcal{T}}^{(n)}\}$ and $\mathcal{V}=\{\bm{\mathcal{V}}^{(1)}, \ldots, \bm{\mathcal{V}}^{(n)}\}$ are called the training and the validation sets, respectively. We wish to perform clustering on the elements of $\mathcal{S}$ in such a way that the groups are associated with global models minimizing the overall forecasting error with respect to the validation set. 
	
	The method we propose is an iterative algorithm having the classical two stages: (i) constructing a prototype for each cluster, usually referred to as centroid and (ii) assigning each series to a specific group. The assignment step often relies on the distance from the series to the prototypes. In this work, we propose to consider global models as prototypes for each group. Specifically, the prototype of the $k$th cluster is a global model which is fitted to the series pertaining to that group.
	
	Assume there are $n_k$ series in the $k$th group $C_k$, i.e., $C_k=\left\{ \bm{X}_{t,k}^{(1)}, \ldots, \bm{X}_{t,k}^{(n_k)} \right\}$, $k=1,\ldots,K$, where the subscript $k$ is used to indicate that the corresponding series belong to cluster $k$. A global model $\mathcal{M}_k$ is fitted in cluster $C_k$ by considering the training periods associated to $\bm{X}_{t,k}^{(j)}$, $j=1,\ldots,n_k$. It is expected that the predictive ability of model $\mathcal{M}_k$ with respect to the series in cluster $C_k$ is better the more related the series in the group are. Note that the set of clusters $\bm C=\{C_1,\ldots,C_K\}$ produce the set of prototypes $\bm{\mathcal{M}}=\{\mathcal{M}_1, \ldots, \mathcal{M}_K\}$.
	
	Once the global models $\mathcal{M}_1, \ldots, \mathcal{M}_K$ have been constructed, each series is assigned to the cluster whose prototype gives rise to the minimal value for a certain error metric by considering its validation period. Specifically, series $\bm{X}_t^{(i)}$, $i=1,\ldots,n$, is assigned to cluster $k'$ such that
	
	\begin{equation}\label{argminerror}
		\begin{split}
			k'=\argmin_{k=1,\ldots,K} d\big(\bm{X}_t^{(i)},\mathcal{M}_k\big),
		\end{split}
	\end{equation}
	
	\noindent where $d(\cdot, \cdot)$ is any function measuring discrepancy between the actual values of  $\bm{X}_t^{(i)}$ and their predictions according to model $\mathcal{M}_k$. For instance, if the mean absolute error (MAE) is considered, then \eqref{argminerror} becomes
	
		\begin{equation}\label{argminmae}
		\begin{split}
			k'=\argmin_{k=1,\ldots,K} d_{\text{MAE}}\big(\bm{X}_t^{(i)},\mathcal{M}_k\big),
		\end{split}
	\end{equation}
	
	 \noindent where $d_{\text{MAE}}\big(\bm{X}_t^{(i)},\mathcal{M}_k\big)=\frac{1}{s(i)}\sum_{j=1}^{s(i)}\big|v_j^{i}-F^{(i)}_{j,k}\big|$ and $F^{(i)}_{j,k}$ is the prediction of $v_j^{i}$ by using the global model $\mathcal{M}_k$. Note that considering the MAE is appropriate in this context, since we are evaluating the forecasting effectiveness of $K$ global models with respect to a single series. Therefore, each assignation is only influenced by the units of the corresponding series so that no scaling issues arise. In fact, the simplicity of the MAE makes it a recommended error metric for assessing accuracy on a single series \cite{hyndman2006another}. Based on previous comments and, unless otherwise stated, we assume that the reassignation rule employed throughout the manuscript is given by \eqref{argminerror}. 
	
	Both steps the computation of prototypes and the reassignation of the series are iterated until convergence or a maximum number of iterations is reached. The corresponding clustering algorithm is described in Algorithm\ref{algorithm1}. Below we provide some remarks concerning the proposed method. \\

	\begin{algorithm}
		\caption{The proposed clustering algorithm based on prediction accuracy of global forecasting models \label{algorithm1}}
		\begin{algorithmic}[1]
			\State Fix $K$, $l$ and $max.iter$%\Comment Number of clusters, fuziness    
			\State Set $iter \, =2$ 
			\State Randomly divide the $n$ series into $K$ clusters
			\State Compute the initial set of $l$-lagged global models $\bm {\mathcal{M}} =\{\mathcal{M}_1, \ldots, \mathcal{M}_K\}=\bm {\mathcal{M}}^{(1)}$
			\Repeat
			\State Set $\bm {\mathcal{M}}_{\text{OLD}}=\bm {\mathcal{M}}^{(iter-1)}$   
			\Comment{Store the current prototypes}
			\State Assign each series to the cluster associated with its nearest prototype according to the rule in \eqref{argminerror}
			\State Compute the new collection of prototypes, $\bm {\mathcal{M}}^{(iter)}$, by fitting a $l$-lagged global model to the training periods of the series in $k$th cluster, $k=1,\ldots,K$. 
			\Comment{Update the set of prototypes}
			\State $iter \, \gets iter \, + 1$
			\Until{ \mbox{ $\bm {\mathcal{M}}= \bm {\mathcal{M}}_{\text{OLD}} \mbox{ or } iter \, = \, max.iter$ } }
            \State Considering the final set of $K$ clusters, construct the final collection of prototypes by fitting a $l$-lagged global model to the training and validation periods of the series in $k$th cluster, $k=1,\ldots,K$.  
		\end{algorithmic}
	\end{algorithm}

	\noindent \textbf{Remark 1} (\textit{Interpretation of the objective function}). Note that the objective function in Algorithm \ref{algorithm1} can be written as 
	
	\begin{equation}\label{of}
		J(\bm C)=\sum_{k=1}^{K}\sum_{{\substack{i=1:\\ \bm X_t^{(i)} \in C_k}}}^{n}d(\bm X_t^{(i)}, \mathcal{M}_k),
	\end{equation}

	\noindent which is a sum of prediction errors with respect to the validation periods. In particular, each series is forecasted by using the global model associated with the cluster it pertains to. In this regard, the value of the objective function returned when Algorithm \ref{algorithm1} stops, say $J_{\text{OPT}}$, can be regarded as the total optimal (minimal) prediction error when $K$ groups are assumed to exist in the dataset. In the same way, the quantity $J_{\text{OPT}}/n$ can be interpreted as the average optimal prediction error. In sum, the objective function of the proposed clustering algorithm is very interpretable from a forecasting perspective. \\
	
	\noindent \textbf{Remark 2} (\textit{Assessment of the predictive accuracy}). Although the quantity $J_{\text{OPT}}/n$ can be seen as the average optimal prediction error (see Remark 1), this value is not an appropriate metric to assess the predictive ability of the resulting global models. In fact, note that the two-step procedure described in Algorithm \ref{algorithm1} attempts to find the partition minimizing the average prediction error with respect to the validation periods. Therefore, $J_{\text{OPT}}/n$ is likely to underestimate the prediction error computed over future periods of the series which are not involved in the optimization process. In this regard, a proper error metric could be obtained through the following steps: \vspace{0.1 cm}
	
	\begin{enumerate}
		\item Given a prediction horizon $h \in \mathbb{N}$, divide each series into two periods. The first period contains all but the last $h$ observations of the series. The second period, referred to as test period, contains the last $h$ observations. For the sake of simplicity, the first periods can be identified with the set $\mathcal{S} = \left\{ \bm{X}_t^{(1)}, \ldots, \bm{X}_t^{(n)} \right\}$ introduced above, whereas the second periods constitute a new set $\mathcal{S}^*= \left\{ \bm{X}_t^{(1)*}, \ldots, \bm{X}_t^{(n)*} \right\}$, where each $\bm{X}_t^{(i)*}=(X_1^{(i)*}, \ldots, X_{h}^{(i)*})$ is a series of length $h$. The set $\mathcal{S}^*$ is called the test set. 
		\item Run Algorithm \ref{algorithm1} using the set $\mathcal{S}$ as input, obtaining the clustering solution.
		\item Given the clustering solution computed in Step 2, and for $k=1,\ldots,K$, fit a $l$-lagged global model to the set of series in the $k$th cluster by considering both training and validation periods. This produces the set of global models $\overline{\bm {\mathcal{M}}} =\{\overline{\mathcal{M}}_1, \ldots, \overline{\mathcal{M}}_K\}$. 
		\item Compute the average prediction error with respect to the test set as
		
		\begin{equation}\label{errormetric}
			\frac{1}{n}\sum_{k=1}^{K}\sum_{{\substack{i=1:\\ \bm X_t^{(i)} \in C_k}}}^{n}d^*\big(\bm{X}_t^{(i)*}, \overline{\mathcal{M}}_k\big),
		\end{equation}
		
		\noindent where $d^*(\cdot, \cdot)$ is any function measuring discrepancy between the actual values of  $\bm{X}_t^{(i)*}$ and their predictions according to model $\overline{\mathcal{M}}_k$. Note that these predictions are computed starting from the series $\bm{X}_t^{(i)}$ and in a recursive manner. As an example, if the MAE is chosen as the error metric, then  \eqref{errormetric} becomes $\frac{1}{n}\sum_{k=1}^{K}\sum_{{\substack{i=1:\\ \bm X_t^{(i)} \in C_k}}}^{n}d^*_{\text{MAE}}\big(\bm{X}_t^{(i)*}, \overline{\mathcal{M}}_k\big)$, with
		
		\begin{equation}\label{errormetricmae}
		d^*_{\text{MAE}}\big(\bm{X}_t^{(i)*}, \overline{\mathcal{M}}_k\big)=\frac{1}{h}\sum_{j=1}^{h}\big|X_j^{(i)*}-\overline{F}_{j,k}^{(i)*}\big|,
		\end{equation}
		
		\noindent where $\overline{F}_{j,k}^{(i)*}$ is the prediction of $X_j^{(i)*}$ according to the global model $\overline{\mathcal{M}}_k$. It is worth highlighting that, if all the time series in the set are recorded in the same scale, then employing the MAE leads to meaningful conclusions. However, if that is not the case, \eqref{errormetricmae} is likely to be a misleading performance measure, since series taking higher values are expected to have a larger impact on the computation of the average prediction error. This issue can be avoided by considering alternative error metrics (see Section \ref{sectionapplication}).

	\end{enumerate}

	\noindent \textbf{Remark 3} (\textit{Numerical behaviour of the algorithm}). The optimisation procedure presented in Algorithm 1 does not guarantee a decreasing in the value of the objective function $J(\bm C)$ from one iteration to the next, as it is the case with other standard clustering methods (e.g., $K$-means). This is due to the fact that new global models are being fitted at each step. Although it is reasonable to expect that the rule in \eqref{argminerror} improves the predictive ability of the global models, this is not always ensured. As a result, undesirable situations can arise in some settings, as the algorithm entering an infinite loop with $J(\bm C)$ showing a continuous increasing-decreasing pattern. These drawbacks can be mitigated by introducing an additional stopping criterion in Algorithm \ref{algorithm1} as follows. Fixed $L \in \mathbb{N}$, the algorithm stops if no improvement in the value of $J(\bm C)$ took place during the last $L$ iterations. In case the algorithm stops due to this rule, the returned clustering solution is the one associated with the minimum value of $J(\bm C)$. \\

 Note that two important input parameters have to be set in advance before executing Algorithm \ref{algorithm1}, namely the number of considered lags to fit the global models ($l$) and the number of clusters ($K$). These two parameters can be easily selected by: (i) running the clustering algorithm in a grid of values for the pair $(l,K)$, and (ii) choosing the combination giving rise to the minimum value of the average error computed with respect to the test set (see Remark 2). In  this way, the optimal pair in terms of predictive effectiveness is selected. The previous procedure is summarized in Algorithm \ref{algorithm2}. Note that, for a fixed $l$, the case $K=1$ corresponds to a global model fitted to all the series, whereas the case $K=n$ corresponds to a local model fitted to each series (local approach).

 \begin{algorithm}
		\caption{A procedure for selecting the number of clusters ($K$) and the number of lags ($l$) \label{algorithm2}}
		\begin{algorithmic}[1]
			\State Consider a grid of values for $(K, l)$, $\mathcal{G}=\{(K_i, l_j): i=1, \ldots, n_K, j=1,\ldots,n_l\}$, with $n_K, n_l \in \mathbb{N}$    
			\State For each $g=(\overline{K}, \overline{l})\in \mathcal{G}$, run Algorithm \ref{algorithm1} by fitting $\overline{l}$-lagged global models in a number of $\overline{K}$ clusters. Compute the average prediction error with respect to the test set as indicated in  \eqref{errormetric} 
			\State Choose the element $g^*=(\overline{K}^*, \overline{l}^*)\in \mathcal{G}$ such that the prediction error is minimal. The associated partition returned by Algorithm \ref{algorithm1} for $g^*$ is the optimal clustering solution  
		\end{algorithmic}
	\end{algorithm}

 Note that the procedure for choosing $K$ and $l$ described in Algorithm \ref{algorithm2} is mainly based on maximizing the predictive accuracy, which is a reasonable and natural rule. In this regard, it is worth remarking that several heuristic criteria are available for the selection of parameter $K$ (e.g., rules based on internal indexes as the Silhouette index), which constitutes an important problem in the clustering literature. According to previous considerations, such criteria are not necessary when carrying out clustering by means of the proposed approach, which constitutes an advantage of CPAGM with respect to alternative techniques for TSC. It is worth highlighting that the class of global models constitutes another important parameter to be selected before running the clustering procedure. 

    \section{Motivating example}\label{sectionmotivatingexample}
    
\begin{figure}
		\centering
		\includegraphics[width=1\textwidth]{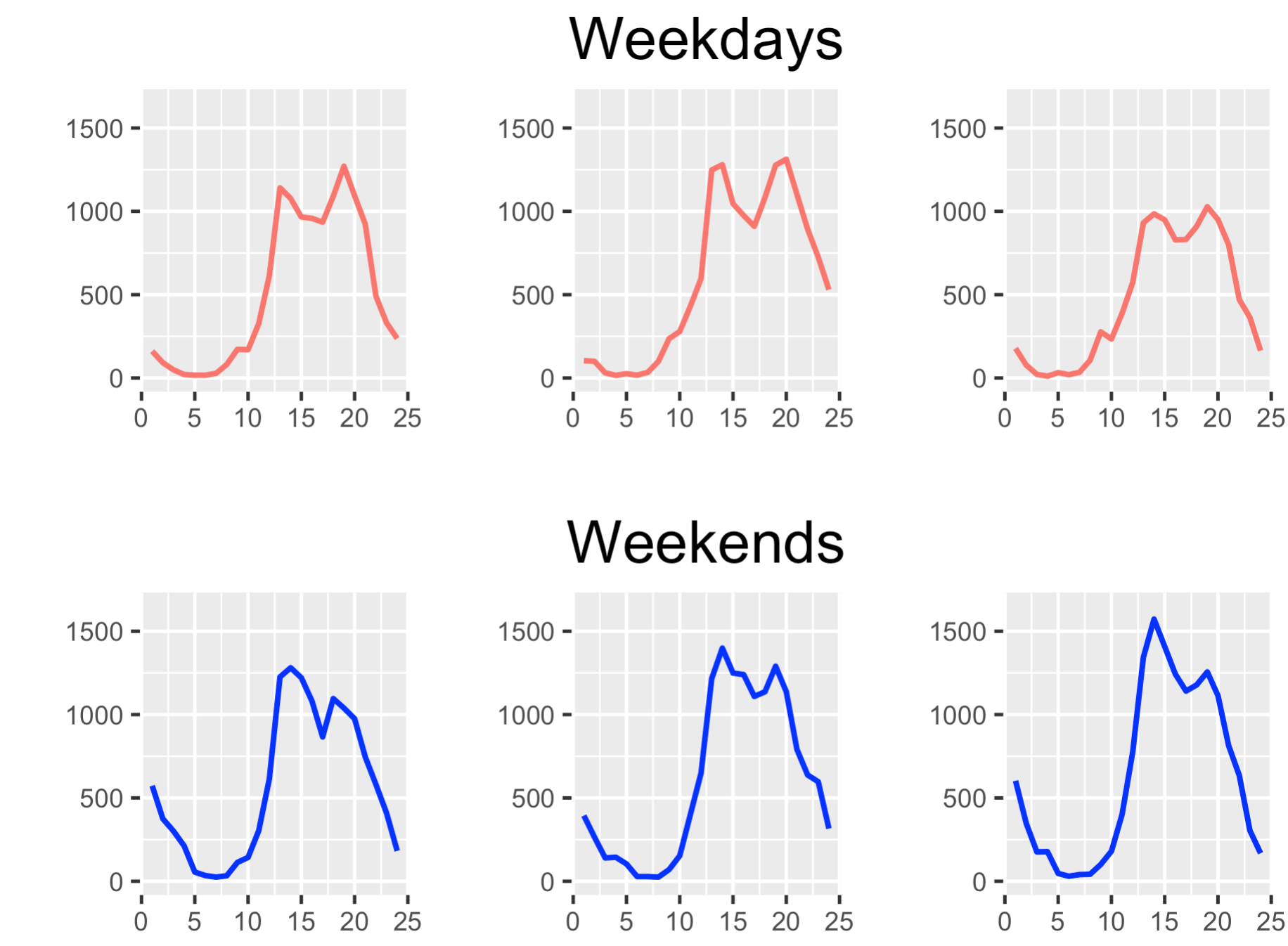}
		\caption{Three series representing business days (top panels) and weekend days (bottom panels) in dataset Chinatown.}
		\label{chinatown}
	\end{figure}

    In order to illustrate the usefulness of the clustering procedure presented in the previous section, we considered a real time series dataset called Chinatown, which pertains to the well-known UCR time series archive\footnote{\href{http://www.timeseriesclassification.com/index.php}{http://www.timeseriesclassification.com/index.php}} \cite{dau2019ucr}. This archive consists of a collection of heterogeneous time series databases which are frequently used to evaluate the accuracy of different machine learning algorithms for temporal data \cite{ding2020novel, bostrom2017binary}, including TSC \cite{zakaria2012clustering, ma2017distance}. Dataset Chinatown includes data recorded by an automated pedestrian counting system located in a specific street of Melbourne, Australia, during the year 2017. Specifically, Chinatown includes 363 time series of length 24, with each series being associated with a particular day of the year and each time observation with a particular hour (e.g., the fifth observation corresponds to 5:00 am). Thus, each series measures the hourly number of pedestrians in the corresponding street. Originally, two classes of series are assumed to exist in dataset Chinatown according to whether the data come from normal business days or weekend days (true partition). The top and bottom panels of Figure \ref{chinatown} contain three series associated with normal and weekend days, respectively. As expected, there are some differences between both groups of series, indicating that the temporal evolution of the number of pedestrians on any given day is clearly influenced by whether or not that day is a normal business day. For instance, it seems that the number of pedestrians reaches its peak during the afternoon on weekends, but during the evening on weekdays. In addition, the number of people just after midnight (e.g., 1 am) is higher on weekends, which is expected. 
	
\begin{figure}
	\centering
	\includegraphics[width=0.8\textwidth]{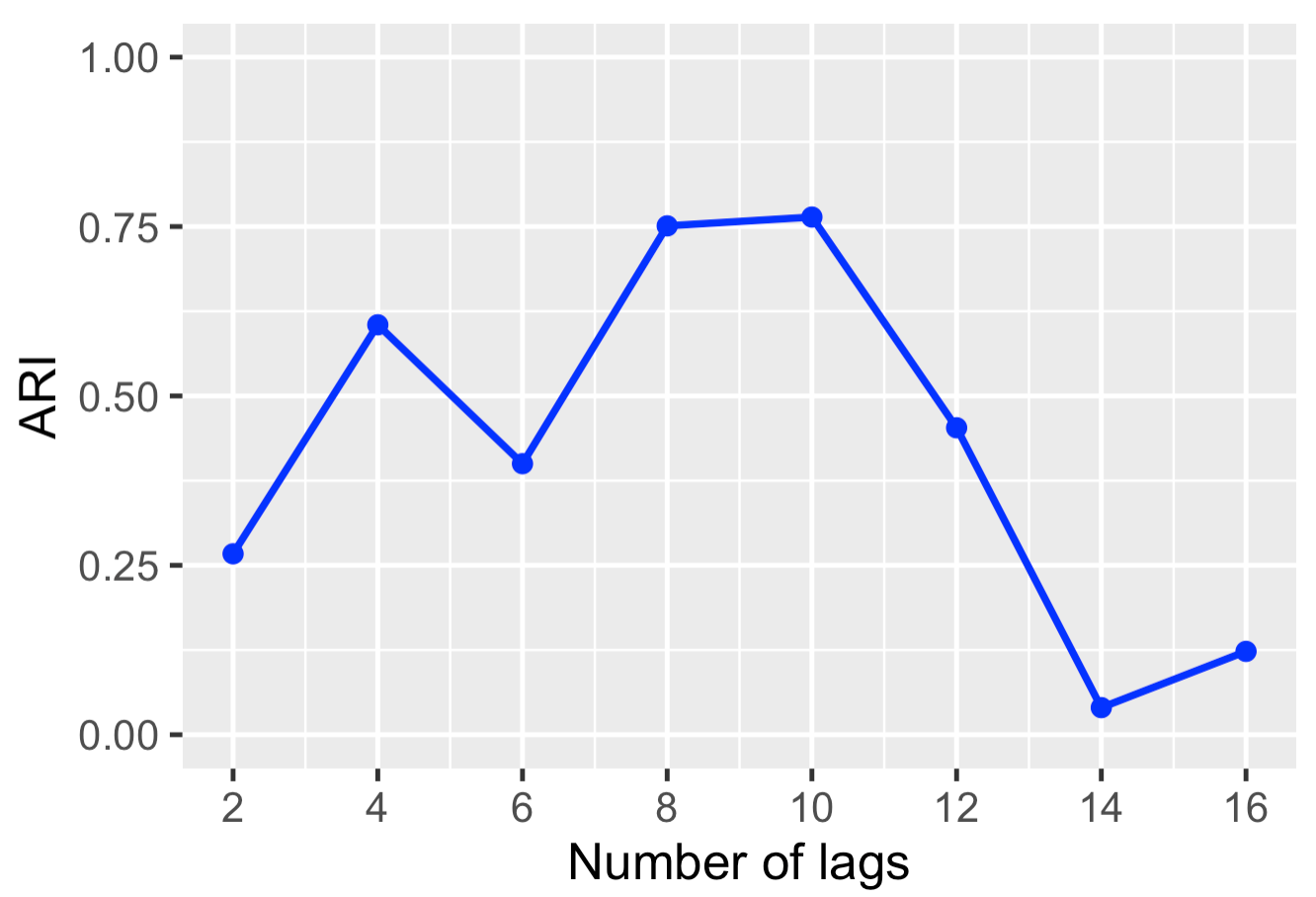}
	\caption{ARI as a function of the number of lags in dataset Chinatown.}
	\label{aril}
\end{figure}

The clustering procedure defined in Algorithm \ref{algorithm1} was applied to the time series in dataset Chinatown by using linear global models fitted by least squares. For the sake of illustration, we considered $K=2$ (as there are two different classes in the original dataset) and run the algorithm for different values of $l$, namely $l \in \{2, 4, \ldots, 16\}$. We constructed a test set by considering the last $h=5$ observations of each series. The training period was set to the first 19 observations of each series, while the validation period was set to observations from $l+1$ to 19. Therefore, the reassignation step in Algorithm \ref{algorithm1} is performed by using the in-sample error (see \eqref{argminerror}). For each value of $l$, the corresponding experimental partition was compared with the true one by considering the adjusted Rand index (ARI) \cite{hubert1985comparing}, which is bounded between $-1$ and $1$. Values of ARI close to $0$ indicate a noninformative clustering solution, while the closer to $1$ the index, the better is the agreement between both partitions. Figure \ref{aril} contains a curve representing the corresponding ARI values as a function of the number of lags used to fit the global models ($l$). Note that high ARI values are achieved for $l=8$ and $l=10$, while the degree of similarity between both partitions decreases when more lags are considered. Specifically, the highest ARI value (namely 0.764) is reached when $l=10$. 

\begin{figure}
	\centering
	\includegraphics[width=0.8\textwidth]{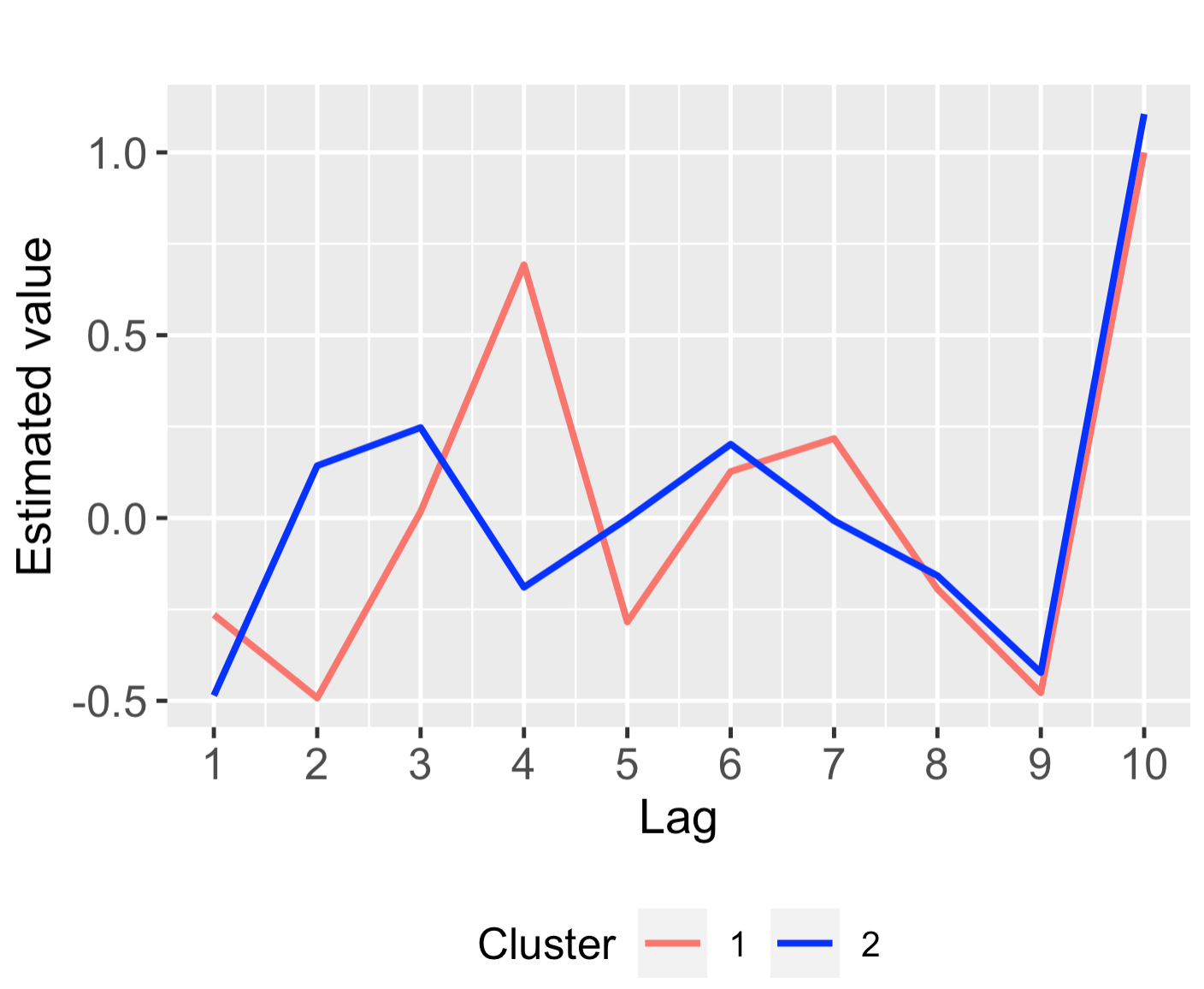}
	\caption{Estimated coefficients for lags 1 to 10 for the global linear models concerning the 2-cluster solution produced by the proposed algorithm ($l=10$) in dataset Chinatown.}
	\label{lagschinatown}
\end{figure}

To gain greater insights into the behavior of the proposed algorithm in dataset Chinatown, we decided to analyze the clustering solution associated with $l=10$. Specifically, we chose to examine the resulting prototypes, i.e., the final global models. In fact, these models characterize the forecasting structures of the different groups, and their analysis can provide a meaningful description of the time series belonging to each cluster. Note that, as linear global models were considered, simple descriptions can be given by providing the corresponding estimated coefficients. Figure \ref{lagschinatown} displays the estimated coefficients for the prototypes of both groups, which were labeled as Clusters 1 and 2. In particular, Cluster 1 contains mostly series associated with normal days, while Cluster 2 includes mainly series associated with weekend days. While the estimated coefficients for lag 8 and beyond are  very similar for both prototypes, there are clear differences at earlier lags. For instance, the estimated coefficients for lags 3, 5 and 7 are close to zero for one of the groups but significantly different from zero for the remaining one, which indicates that both prototypes show a different behavior. Additionally, the estimates for the intercepts of global models associated with Clusters 1 and 2 are 447.82 and 499.06, respectively, thus suggesting that the number of pedestrians is higher during the weekends. 

\begin{figure}
	\centering
	\includegraphics[width=0.8\textwidth]{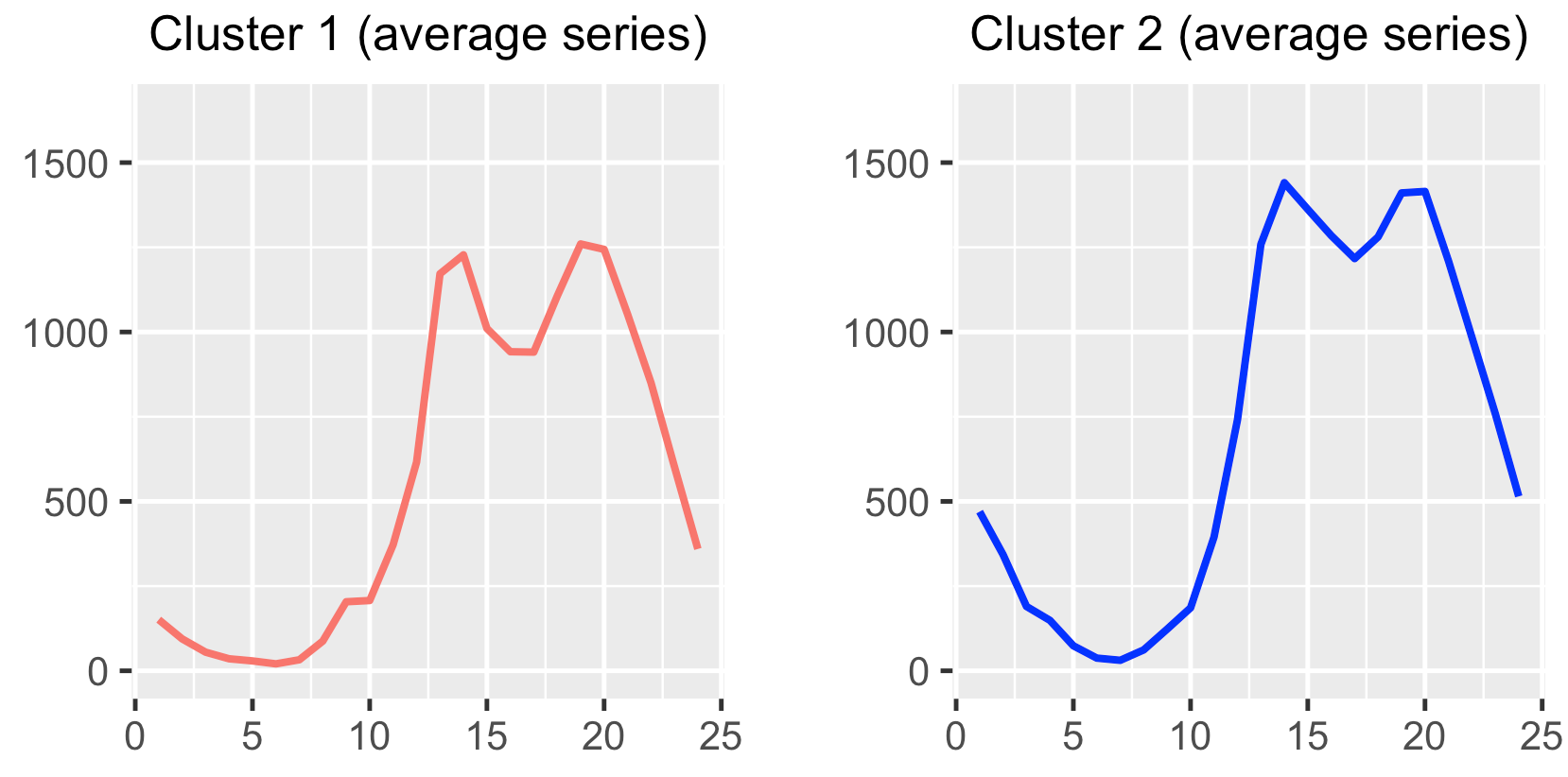}
	\caption{Average time series for the 2-cluster solution defined by the prototypes in Figure \ref{lagschinatown}.}
	\label{avgseries}
\end{figure}

For illustrative purposes, we constructed the average time series associated with the clustering solution defined by the prototypes in Figure \ref{aril} ($l=10$). That is, for each one of the groups, we calculated the average value of all the time series belonging to that group at each time point. The resulting series for Clusters 1 and 2 are displayed in the left and right panels of Figure \ref{avgseries}, respectively. Note that these plots are coherent with the series in Figure \ref{chinatown} and with previous comments. In fact, the average series for the weekend group (right panel) takes higher values (in particular at early hours) than the average series for the weekday group (left panel). Moreover, the former series indicates a peak in the number of pedestrians just after noon, while this peak does not happen until the evening according to the latter series. Previous analyses suggest that the proposed algorithm is able to clearly identify the weekday-weekend pattern of dataset Chinatown when a suitable number of lags is considered. 

Although the main goal of the proposed method is to detect the different forecasting patterns existing in a given dataset, an interesting side effect of Algorithm \ref{algorithm1} is that the resulting prototypes are expected to improve the prediction accuracy of a single global model when $K$ groups of series exist. In fact, by considering the clustering partition associated with $l=10$, the average MAE with respect to the test set (see Remark 2) is 370.42, while this quantity takes the value of 535.08 when only one global model is fitted to all the series ($K=1$). Hence, splitting the dataset into two clusters results in substantially better predictions. Similar results to the ones provided above  are obtained with different values of $h$ are considered. 

In short, this section showed an example where the forecasting patterns detected by the proposed algorithm are associated with highly interpretable classes (namely weekday and weekend days) of a dataset which is frequently used in the TSC literature. 

	\section{Simulation study}\label{sectionsimulationstudy}
	
	In this section we perform several simulations with the aim of assessing the performance of the proposed approach in different scenarios. Firstly we describe the simulation mechanism, then we explain how the evaluation of the method was done and, afterwards, we show the results of the simulation study. Finally, we carry out a set of additional experiments related to the numerical behaviour of the algorithm, the selection of some hyperparameters, or the consideration of complex models, among others. 
	
	\subsection{Experimental design}\label{subsectionexperimentaldessign}
	
	Two unsupervised classification setups involving linear processes were considered, namely clustering of (i) short memory processes and (ii) long memory processes. In this way, the proposed method was analysed under very dissimilar serial dependence structures. Both settings contain three different generating processes. The specific scenarios and the generating models are given below. \\
	
	\noindent \textbf{Scenario 1}. Let $\{X_t\}_{t \in \mathbb{Z}}$ be a stochastic process following the AR($p$)-type recursion given by
	
	\begin{equation}\label{arp}
		X_t=\sum_{i=1}^{p}\varphi_iX_{t-i}+\epsilon_t,
	\end{equation}

	\noindent where $\varphi_1,\ldots, \varphi_p$ are real numbers verifying the corresponding stationarity condition and $\{\epsilon_t\}_{t \in \mathbb{Z}}$ is a process formed by independent variables following the standard normal distribution. We fix $p=4$. The vector of coefficients $\bm \varphi_4=(\varphi_1, \varphi_2, \varphi_3, \varphi_4)$ is set as indicated below. \\

	\noindent Process 1: $\bm \varphi_4=(0.1, 0.2, -0.4, 0.3)$. \\
	\noindent Process 2: $\bm \varphi_4=(0.2, -0.5, 0.3, -0.3)$.\\
	\noindent Process 3: $\bm \varphi_4=(-0.3, 0.4, 0.6, -0.2)$. \\
	
	\noindent \textbf{Scenario 2}. Consider the AR($p$) process given in \eqref{arp}. We fix $p=12$. The vector of coefficients $\bm \varphi_{12}=(\varphi_1, \varphi_2, \ldots, \varphi_{12})$ is set  as \\
	
	\noindent {\small $(0.9, -0.5, -0.3, 0.3, 0.1, -0.3, 0.2, -0.3, 0.5, -0.5, 0.3, -0.3)$}, \\
	\noindent {\small $(0.2, 0.3, -0.2, -0.2, 0.4, 0.2, -0.1, 0.2, 0.1, -0.2, -0.3, 0.5)$},\\
	\noindent  {\small $(-0.3, -0.1, 0.3, -0.1, -0.2, -0.1, -0.4, -0.2, -0.3, 0.4, 0.1, 0.2)$}, \\
	
	\noindent for Processes 1, 2 and 3, respectively.

	The simulation study was carried out as follows. For each scenario, $N$ time series of length $T$ were generated from each process. Several values of $N$ and $T$ were taken into account to analyse the effect of those parameters (see Section \ref{subsectionresultsdiscussion}). The test set was constructed by considering the last $h=2l_{\text{SIG}}$ observations of each series, where $l_{\text{SIG}}$ is the number of significant lags existing in each scenario (e.g., $l_{\text{SIG}}=4$ in Scenario 1). The training period was set to the first $(T-h)$ observations of each series. The validation period was set to observations from $(l+1)$ to $(T-h)$. Note that this choice implies that the reassignation step in Algorithm \ref{algorithm1} is carried out by considering the in-sample error (see \eqref{argminerror}). The simulation procedure was repeated 200 times for each pair $(T, N)$. 
	
	\subsection{Alternative approaches and assessment criteria}\label{subsectionalternativeapproaches}
	
	To throw light on the behaviour of the proposed algorithm, which we will refer to as \textit{Clustering based on Prediction Accuracy of Global Models} (CPAGM), we decided to compare it with the alternative approaches described below.
	
	\begin{itemize}
		\item \textit{Local Models} (LM). Specifically, a local model (e.g., an AR model) is fitted to each series in the collection (by jointly considering training and validation periods) and used to obtain the predictions with respect to the test period. In this way, each local model gives rise to an error metric measuring its predictive accuracy. The average of these quantities can be seen as the overall error associated with the LM approach. Note that the LM method was already used by \cite{montero2021principles} to show the benefits of global models for forecasting purposes. 
		\item   \textit{Global Models by considering an Arbitrary Partition} (GMAP). This procedure is based on 2 steps: (i) the original set of series $\mathcal{S}$ is randomly partitioned into $K$ groups and (ii) for each group, a global model is fitted by considering the series pertaining to that cluster. The assessment task is carried out as indicated in Step 4 of Remark 2. It is worth highlighting that global models fitted to random groups of series have been shown to improve the predictive accuracy of one global model fitted to all the series in some datasets (see, e.g., Figure 4 in \cite{montero2021principles}). The approach GMAP can be seen as a meaningful benchmark for the proposed method, since it is expected that the groups produced by Algorithm \ref{algorithm1} improve the forecasting effectiveness of the corresponding global models in comparison with a random partition. 
		\item \textit{Global models by considering Feature-Based Clustering} (GMFBC). Particularly, the technique proposed by \cite{bandara2020forecasting}, which relies on two steps: (i) the original collection of series is splitted into $K$ groups by using a clustering algorithm based on the feature extraction procedure described in \cite{hyndman2015large} and (ii) $K$ global models are constructed according to the resulting partition. This approach is evaluated in a similar way that GMAP. Note that, like CPAGM, GMFBC also tries to exploit the notion of similarity between time series in order to minimize the overall prediction error. However, GMFBC considers a specific clustering algorithm before fitting the global models, while CPAGM iterates until achieving the optimal clustering partition in terms of forecasting effectiveness. 
	\end{itemize}
	
	In the simulations, the number of clusters was set to $K=3$, since both scenarios contain 3 different generating processes. For approaches CPAGM, GMAP and GMFBC, the number of lags $l$ to fit the global models was set to $l=l_{\text{SIG}}$. The considered global models were standard linear regression models adjusted by least squares. As for the method LM, a linear local model was fitted to each series by using the function \textit{auto.arima()} in the \textbf{forecast} R package \cite{hyndman2015forecasting}. Model selection was performed by means of AICc criterion. Note that classical linear models are important as a benchmark because they do not include any advanced machine learning technique and overlap the model class with ARIMA model (a common local approach). Therefore, they are ideal to isolate the effect of globality \cite{montero2021principles} and to analyze the advantages of splitting the dataset into different groups according to Algorithm \ref{algorithm1}. 
	
	The quality of the procedures was evaluated by comparing the clustering solution given by the algorithms with the true partition, usually referred to as ground truth, which is defined in each scenario by the corresponding underlying processes. Approaches CPAGM and GMFBC automatically provide a clustering partition. For method LM, each series was first described by means of the vector of estimated model coefficients returned by \textit{auto.arima()} function (when necessary, the vectors were padded with zeros until reaching the length of the longest vector). Next, a standard $K$-means algorithm was executed by using these feature vectors as input. Similar clustering methods were already employed by \cite{piccolo1990distance} and \cite{maharaj1999comparison}, the latter in the multivariate setting. Experimental and true partitions were compared by considering the ARI. 
	
	The predictive accuracy of methods CPAGM, GMAP and GMFBC was assessed by recording the average MAE as indicated in \eqref{errormetricmae}. The MAE associated with each local model computed with respect to the test set was stored for LM and the average of those quantities was calculated as the error metric. Note that, since all series within a given scenario have the same numerical scale, the MAE is a proper measure to evaluate the overall prediction error. 
	
	In each simulation trial and given a pair $(T, N)$, the proposed technique CPAGM was executed 5 times and the partition associated with the minimum value of $J_\text{OPT}$ (see Remark 1) was stored. This way, we tried to avoid the well-known issue of local optima related to iterative clustering procedures. A similar strategy was employed for the remaining approaches. The overall MAE produced by GMAP was approximated via Monte Carlo (i.e., by considering several random partitions). 
	
	\subsection{Results and discussion}\label{subsectionresultsdiscussion}
	
	Average values of ARI attained by the different techniques in Scenario 1 are provided in Table \ref{tableari1}. In order to perform rigorous comparisons, pairwise paired $t$-tests were carried out by taking into account the 200 simulation trials. In all cases, the alternative hypotheses stated that the mean ARI value of a given method is greater than the mean ARI value of its counterpart. Bonferroni corrections were applied to the set of $p$-values associated with each value of $T$. An asterisk was incorporated in Table  \ref{tableari1} if the corresponding method resulted significantly more effective than the remaining ones for a significance level 0.01.

	\begin{table}
		\centering
		\resizebox{6cm}{!}{\begin{tabular}{cccc} 
			\hline $(T,N)$ & LM & CPAGM & GMFBC \\   \hline 
			(20, 5) & 0.027 & \textbf{0.352}$^*$ & 0.094 \\   
			(20, 10) & 0.032 & \textbf{0.459}$^*$ & 0.090 \\   
			(20, 20) & 0.029 & \textbf{0.556}$^*$ & 0.092 \\   
			(20, 50) & 0.026 & \textbf{0.612}$^*$ & 0.076 \\  \hline   
			(50, 5) & 0.305 & \textbf{0.914}$^*$ & 0.243 \\   
			(50, 10) & 0.336 & \textbf{0.956}$^*$ & 0.222 \\   
			(50, 20) & 0.331 & \textbf{0.988}$^*$ & 0.216 \\   
			(50, 50) & 0.331 & \textbf{0.981}$^*$ & 0.195 \\ \hline
			(100, 5) & 0.747 & \textbf{0.946}$^*$ & 0.379 \\  
			(100, 10) & 0.740 & \textbf{0.954}$^*$ & 0.380 \\   
			(100, 20) & 0.743 & \textbf{0.961}$^*$ & 0.334 \\  
			(100, 50) & 0.740 & \textbf{0.956}$^*$ & 0.311 \\ \hline 
			(200, 5) & 0.876 & \textbf{0.906}& 0.581 \\ 
			(200, 10) & 0.854 & \textbf{0.919}$^*$ & 0.561 \\  
			(200, 20) & 0.820 & \textbf{0.921}$^*$ & 0.516 \\   
			(200, 50) & 0.800 & \textbf{0.926}$^*$ & 0.488  \\  \hline 
			(400, 5) & 0.897 & \textbf{0.908} & 0.719 \\   
			(400, 10) & 0.848 & \textbf{0.900}$^*$ & 0.725 \\  
			(400, 20) & 0.877 & \textbf{0.881} & 0.732 \\  
			(400, 50) & 0.803 &\textbf{0.872}$^*$ & 0.726 \\    \hline \\
		\end{tabular}}
		\caption{Average ARI in Scenario 1. For each pair $(T, N)$, the best result is shown in bold. An asterisk indicates that a given method is significantly better than the rest at level $\alpha=0.01$.}
		\label{tableari1}
	\end{table}

	According to Table \ref{tableari1}, the proposed method CPAGM achieved significantly greater ARI values than the alternative approaches in most cases. The only exceptions were $(T,N)=(200,5)$, $(T,N)=(400,5)$ and $(T,N)=(400,20)$, where CPAGM and LM showed a similar performance. What happens here is that, as long series are considered, the model coefficients are very accurately estimated via the local approach and then the clustering partition returned by LM is quite similar to the ground truth. An increasing in the number of series per cluster was clearly beneficial for the proposed method when short series were considered $(T \in \{20, 50\})$, but it had little impact when $T>50$. In some way, considering more series per cluster has a similar effect on CPAGM than increasing the series length, since both phenomena result in a better estimation of the global models. The approach GMFBC showed a steady improvement when increasing the series length, but it was still far from a perfect partition for $T=400$.

	Average results for Scenario 2 concerning ARI are displayed in Table \ref{tableari2}. The proposed approach showed a similar behaviour than in Scenario 1 in terms of clustering effectiveness, but the differences with respect to the remaining techniques were more marked in Scenario 2. The long memory patterns exhibited by the processes of this scenario negatively affected both methods LM and GMFBC. In fact, the local approach was not able to show the same performance than CPAGM even when very long series ($T=1000$) were considered. In short, the iterative procedure of Algorithm \ref{algorithm1} takes advantage of the excellent accuracy of global models to properly estimate the complex forecasting patterns existing in the long memory processes of Scenario 2.  
	
	\begin{table}
		\centering
		\resizebox{6cm}{!}{\begin{tabular}{cccc}  \hline 		
			$(T,N)$ & LM & CPAGM & GMFBC \\   \hline 
			$(50, 5)$  & 0.243 & \textbf{0.584}$^*$ & 0.238 \\   
			$(50, 10)$  & 0.259 & \textbf{0.853}$^*$ & 0.222 \\  
			$(50, 20)$  & 0.250 & \textbf{0.956}$^*$ & 0.219 \\  
			$(50, 50)$  & 0.256 & \textbf{0.980}$^*$ & 0.205 \\ \hline
			$(100, 5)$  & 0.386 & \textbf{0.933}$^*$ & 0.278 \\   
			$(100, 10)$  & 0.387 & \textbf{0.937}$^*$ & 0.274 \\  
			$(100, 20)$  & 0.410 & \textbf{0.979}$^*$ & 0.277 \\ 
			$(100, 50)$  & 0.412 & \textbf{0.986}$^*$ & 0.286 \\  \hline
			$(200, 5)$  & 0.453 & \textbf{0.907}$^*$ & 0.302 \\ 
			$(200, 10)$  & 0.478 &\textbf{0.937}$^*$ & 0.317 \\  
			$(200, 20)$  & 0.468 & \textbf{0.959}$^*$ & 0.306 \\  
			$(200, 50)$  & 0.477 & \textbf{0.972}$^*$ & 0.303 \\  \hline
			$(400, 5)$  & 0.517 & \textbf{0.898}$^*$ & 0.383 \\ 
			$(400, 10)$  & 0.510 & \textbf{0.918}$^*$ & 0.382 \\ 
			$(400, 20)$ & 0.507 & \textbf{0.926}$^*$ & 0.368 \\ 
			$(400, 50)$ & 0.487 & \textbf{0.921}$^*$ & 0.365 \\   \hline
			$(1000, 5)$ & 0.571 & \textbf{0.846}$^*$ & 0.497 \\  
			$(1000, 10)$ & 0.556 & \textbf{0.841}$^*$ & 0.456 \\  
			$(1000, 20)$ & 0.552 & \textbf{0.867}$^*$ & 0.453 \\  
			$(1000, 50)$ & 0.532 & \textbf{0.877}$^*$ & 0.457 \\    \hline \\
		\end{tabular}}
		\caption{Average ARI in Scenario 2. For each pair $(T, N)$, the best result is shown in bold. An asterisk indicates that a given method is significantly better than the rest at level $\alpha=0.01$.}
		\label{tableari2}
	\end{table}
	
	Average results in terms of MAE for Scenarios 1 and 2 are given in Tables \ref{tablemae1} and \ref{tablemae2} in the Appendix, respectively, where a discussion of the performance of the different approaches is also provided. In short, the proposed method significantly outperforms the alternative techniques in most cases, and the differences are particularly pronounced in Scenario 2. 
	
	\subsection{Additional analyses}
	
	This section shows some additional analyses which complement the simulations presented above.

    \subsubsection{Noisy scenarios}\label{subsubsectionnoisyscenario}

    The previous simulations considered scenarios with well-defined clusters given by three types of autoregressive processes. Specifically, the time series belonging to a given group were generated by the same stochastic process. Although this is a reasonable simulation mechanism, it is also interesting to study the behavior of the different methods when some degree of uncertainty exist in the underlying processes. To this aim, we considered a slightly modified version of Scenario 2 by incorporating some amount of noise in the corresponding model coefficients. Particularly, series 1 to $N$ within a given group were simulated from an autoregressive process with vector of coefficients $u_1\boldsymbol{\varphi}_{12}, \ldots, u_N\boldsymbol{\varphi}_{12}$, where $\boldsymbol{\varphi}_{12}$ is the vector of coefficients associated with the corresponding group (see Section \ref{subsectionexperimentaldessign}) and $u_1,\ldots,u_N$ are independent random variables following a uniform distribution in the interval $(0.8, 1)$. Note that, according to previous considerations, each group of series in this new scenario shows a moderate level of variability in terms of generating structures, thus making the clustering task more challenging. The proposed approach and the alternative methods were assessed in this additional setting by following the same steps as in Scenario 2.

    Results in terms of clustering effectiveness for the noisy scenario are given in Table \ref{tablearinoisy2}. Scores in Table \ref{tablearinoisy2} are rather similar to the ones in Table \ref{tableari2}, with method CPAGM significantly outperforming the alternative techbiques in all settings. Note that the clustering accuracy of the former approach does not get negatively affected by the noisy coefficients of the different generating processes. Thus, a moderate amount of uncertainty is not enough to prevent the iterative procedure in Algorithm \ref{algorithm1} from grouping the time series according to the different forecasting structures. It is worth highlighting that the feature-based approach GMFBC substantially decreases its clustering effectiveness with respect to the original Scenario 2, thus indicating that the introduced noise considerably corrupts the estimation of the corresponding statistical quantities. 
    
    \begin{table}
		\centering
		\resizebox{6cm}{!}{\begin{tabular}{cccc}  \hline 		
			$(T,N)$ & LM & CPAGM & GMFBC \\   \hline 
			$(50, 5)$  & 0.180 & \textbf{0.374}$^*$ & 0.094 \\  
			$(50, 10)$  & 0.160 & \textbf{0.732}$^*$ & 0.082 \\  
			$(50, 20)$  & 0.163 & \textbf{0.893}$^*$ & 0.069 \\  
			$(50, 50)$  & 0.160 & \textbf{0.941}$^*$ & 0.051 \\ \hline 
			$(100, 5)$  & 0.343 & \textbf{0.922}$^*$ & 0.140 \\   
			$(100, 10)$  & 0.321 & \textbf{0.978}$^*$ & 0.115 \\  
			$(100, 20)$  & 0.369 & \textbf{0.967}$^*$ & 0.112 \\ 
			$(100, 50)$  & 0.364 & \textbf{0.976}$^*$ & 0.090 \\  \hline
			$(200, 5)$  & 0.461 & \textbf{0.952}$^*$ & 0.180 \\ 
			$(200, 10)$  & 0.497 & \textbf{0.952}$^*$ & 0.149 \\  
			$(200, 20)$  & 0.503 & \textbf{0.957}$^*$ & 0.146 \\  
			$(200, 50)$  & 0.516 & \textbf{0.973}$^*$ & 0.138 \\  \hline
			$(400, 5)$  & 0.572 & \textbf{0.915}$^*$ & 0.227 \\ 
			$(400, 10)$  & 0.593 & \textbf{0.936}$^*$ & 0.190 \\ 
			$(400, 20)$ & 0.572 & \textbf{0.931}$^*$ & 0.183 \\ 
			$(400, 50)$ & 0.575 & \textbf{0.946}$^*$ & 0.180 \\   \hline
			$(1000, 5)$ & 0.681 & \textbf{0.934}$^*$ & 0.248 \\  
			$(1000, 10)$ & 0.687 & \textbf{0.897}$^*$ & 0.227 \\  
			$(1000, 20)$ & 0.687 & \textbf{0.912}$^*$ & 0.217 \\  
			$(1000, 50)$ & 0.702 & \textbf{0.920}$^*$ & 0.214 \\    \hline \\
		\end{tabular}}
		\caption{Average ARI in Scenario 2 with noisy coefficients. For each pair $(T, N)$, the best result is shown in bold. An asterisk indicates that a given method is significantly better than the rest at level $\alpha=0.01$.}
		\label{tablearinoisy2}
	\end{table}
	
	Results in terms of predictive accuracy are provided in Table \ref{tablemaenoisy2} in the Appendix.  In short, the corresponding values indicate that method CPAGM outperforms the remaining techniques in most settings, but the differences in terms of MAE are less marked than in the original Scenario 2. 

    In sum, the previous analysis corroborates the excellent performance of the proposed algorithm even when a moderate amount of noise exists in the generating processes defining the different clusters. Note that this is a great property of CPAGM, since the assumption of clear, well-separated clusters is often not fulfilled in real time series datasets.  
    
	\subsubsection{Selection of $K$ and $l$}

	Note that, in the simulation study of Sections \ref{subsectionexperimentaldessign}, \ref{subsectionalternativeapproaches} and \ref{subsectionresultsdiscussion}, the true values of $K$ and $l$ were given as input to the proposed clustering algorithm. However, the optimal values of these parameters are usually unknown in practice. For this reason, an automatic criterion to perform parameter selection was provided in Algorithm \ref{algorithm2}. In order to study the behaviour of that procedure in practice, we considered Scenario 1 with $(T,N)=(100,5)$. Training and validation sets were the same as in original Scenario 1. Former test periods were split into two parts formed by the first and the last $l_{\text{SIG}}=4$ observations, respectively, giving rise to the corresponding test sets. The first test set was purely used for parameter selection (see the second step above), whereas the second test set was employed for evaluation purposes. 
	
	The procedure described above was run by considering the grid  $\mathcal{G}=\{(K, l):K=1, 2, \ldots, 6,  l=1,2, 3, 4\}$. The average MAE with respect to the first test set was calculated and the pair giving rise to the minimum value of this quantity was selected as the optimal one. The simulation mechanism was repeated 200 times. 
	
	Table \ref{tablekl} shows the percentage of times that each pair $(K,l)$ was chosen. The true combination $(K,l)=(3,4)$ was selected 37\% of the time. The procedure properly detected the correct value of $l$ most of the trials, but identifying the real value of $K$ was more challenging. Particularly, the combinations $(4,4)$, $(5,4)$ and $(6,4)$ were selected with high frequency. It is worth remarking that, although theoretically these pairs could be considered a wrong choice, they are often associated with situations in which: (i) the clustering solution ends up with 3 clusters even though a value $K>3$ is given as input parameter or (ii) the clustering algorithm correctly identifies the three real clusters but in turn divides some of them into further subgroups.

	\begin{table}
		\centering 
		\resizebox{5.5cm}{!}{\begin{tabular}{ccccccc} \hline
			$l$\textbackslash{}$K$ & 1 & 2   & 3   & 4 & 5 & 6    \\ \hline 
			1                  & 0 & 0   & 0 & 0 & 0 & 0 \\
			2                  & 0 & 0   & 0   & 1.5 & 0 & 0  \\
			3                  & 0.5 & 1   & 1   & 1 & 0.5 & 0.5   \\
			4                  & 0.5 & 1.5 & 37   & 13.5 & 23.5 & 18  \\ \hline \\
		\end{tabular}}
		\caption{Percentage of times that each pair $(K,l)$ was selected as the optimal one. The values $T=100$ and $N=5$ were considered.}
		\label{tablekl}
	\end{table}
	
	To analyse to what extent the selection of pairs $(K,4)$ with $K>3$ is appropriate, we computed the average MAE and ARI of such pairs (being the former measure calculated with respect to the second test set) and compared them with the average MAE and ARI associated with the optimal pair. Table \ref{tableklmae} displays the corresponding quantities along with the average MAE and ARI corresponding to the LM approach. Pairs $(K,4)$ with $K \in \{4,5,6\}$ exhibit a similar MAE value than the true pair, which corroborates that those pairs produce clustering partitions as good as the optimal one in terms of forecasting effectiveness. The average MAE associated with local models is significantly higher. All values of the ARI indicate a partition quite close to the ground truth.  
	
	\begin{table}
		\centering 
		\begin{tabular}{ccc} \hline
			Pair $(K,l)$ & Average MAE  &  Average ARI   \\ \hline 
			$(3,4)$                  &  0.8551 & 0.9855  \\
			$(4,4)$                  &   0.8737 & 0.8887 \\
			$(5,4)$                 &  0.8632 &  0.8595 \\ 
			$(6,4)$                  &   0.8505 & 0.8092\\  \hline 
			LM                 &    0.9400 & 0.8427\\  \hline \\
		\end{tabular} 
		\caption{Average MAE (with respect to the second test set) and ARI for several pairs $(K,l)$ and the LM approach. The values $T=100$ and $N=5$ were considered.}
		\label{tableklmae}
	\end{table}
	
	In short, a proper combination of both parameters was selected more than 90\% of the time via the proposed procedure. The numerical experiment was repeated by considering $N>5$ and the optimal pair was selected almost 100\% of the trials. As a last remark, it is worth noting that we did not consider values of $l>4$ in the grid because this often results in global models with estimated coefficients above the $4$th lag being close to zero, which implies that they are virtually equivalent to a $4$-lagged global model. 
	
	\subsubsection{Analysing the iterative behaviour of the proposed method}
	
	The simulations of Sections \ref{subsectionexperimentaldessign}, \ref{subsectionalternativeapproaches} and \ref{subsectionresultsdiscussion} evaluate the performance of the proposed method but without analysing the iterative process described in Algorithm \ref{algorithm1}. However, it is important to assess how the clustering and the predictive accuracy fluctuate from one iteration to the next. To this aim, we executed Algorithm \ref{algorithm1} in a specific setting, namely Scenario 2 with $(T,N)=(100, 20)$, and recorded, in each iteration: (i) the clustering partition, $\bm C$, (ii) the average prediction error with respect to the validation set, $J(\bm C)/n$ and (iii) the average MAE with respect to the test set (see \eqref{errormetricmae}). The numerical experiment described above was repeated 1000 times.

\begin{figure}[ht]
		\centering
		\includegraphics[width=0.7\textwidth]{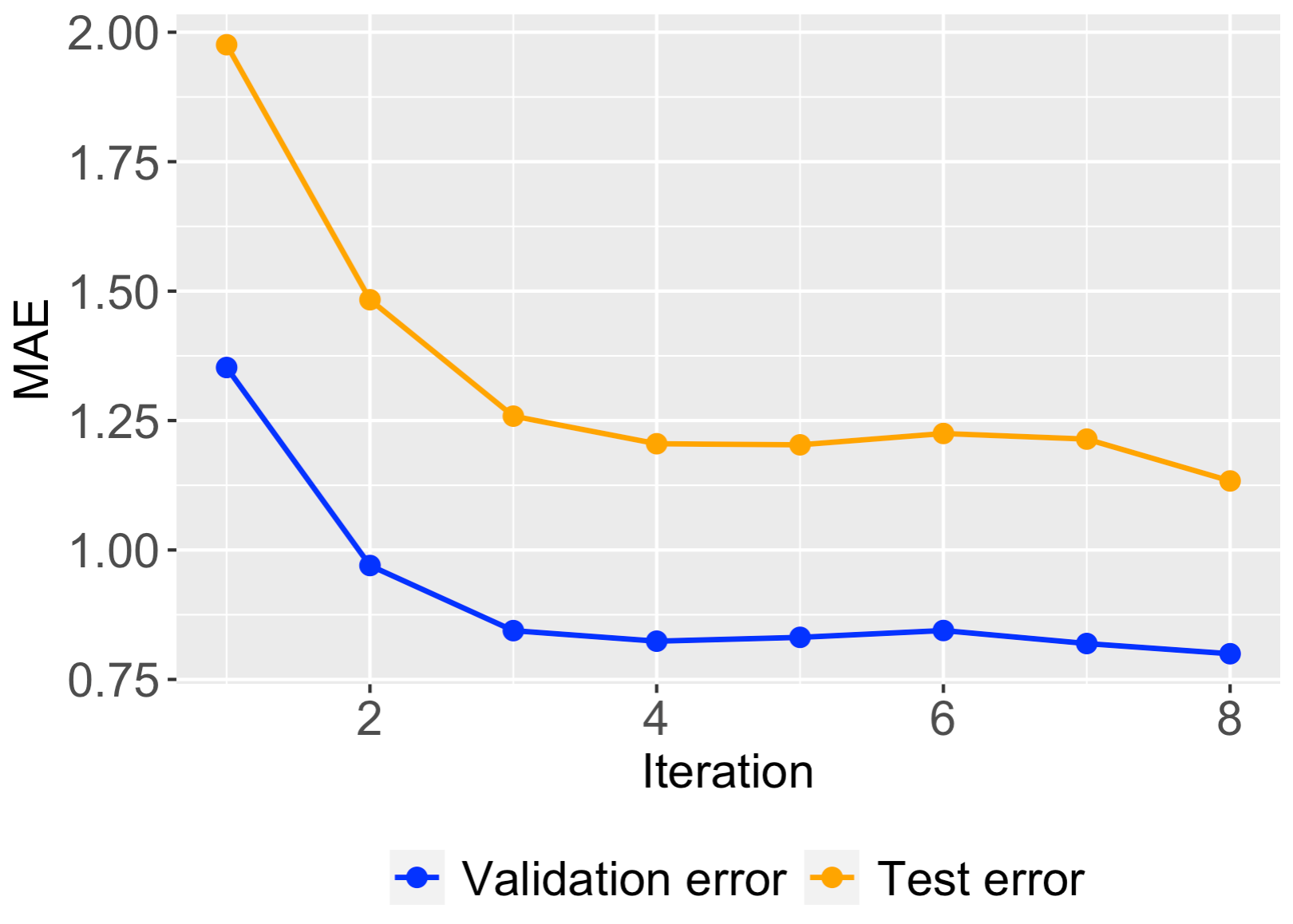}
		\caption{Average MAE with respect to the validation set (blue curve) and test set (orange curve) as a function of the iteration. Scenario 2 with $(T,N)=(100, 20)$ was considered.}
		\label{image_iterations}
	\end{figure}

	The average and maximum number of iterations in the simulation procedure were 3.602 and 8, respectively. Figure \ref{image_iterations} contains two curves displaying the average prediction error (MAE) with respect to the validation (blue colour) and test (orange colour) sets as a function of the specific iteration. Given the $j$th iteration, only those trials in which the clustering algorithm stopped at or after the $j$th iteration were considered to construct the curves in Figure \ref{image_iterations}. Table \ref{tableiterations} shows the specific values of the points depicted in Figure \ref{image_iterations}. The last column includes the average ARI computed by considering the clustering partition associated with each iteration.

	\begin{table}
		\centering 
		\resizebox{7cm}{!}{\begin{tabular}{cccc}
			\hline
			Iteration        & Validation error     & Test error     &  ARI          \\ \hline
			1 & 1.354  & 1.976 & -0.001 \\
			2 & 0.970  & 1.483 & 0.537 \\ 
			3 & 0.844 &  1.259 & 0.848  \\ 
			4  & 0.824  & 1.205 & 0.911 \\
			5 & 0.831  & 1.203 & 0.902 \\
			6 & 0.844  & 1.225 & 0.871 \\
			7 & 0.819  & 1.214 & 0.907 \\
			8 & 0.800  & 1.133 & 1.000 \\ \hline  \\
		\end{tabular}}
		\caption{Average MAE with respect to the validation and test sets and ARI for the corresponding partition, as a function of the iteration. Scenario 2 with $(T,N)=(100, 20)$ was considered.}
		\label{tableiterations}
	\end{table}

	It is clear from Figure \ref{image_iterations} and Table \ref{tableiterations} that the quantity $J(\bm C)/n$ decreases the most during the first three iterations and gets stabilized afterwards. The curve indicating average error with respect to the test set exhibits a similar pattern, implying that a drop in the validation error is accompanied by a similar decrease in the test error. However, the latter curve takes higher values, which corroborates that $J_{\text{OPT}}/n$ underestimates the real forecasting error of the method (see Remark 2).

	Average values of ARI index also improve substantially during the first three iterations. However, after the $3$rd iteration, clustering solutions are always rather similar to the ground truth. Interestingly, the maximum value of 1 (associated with a perfect identification of the underlying partition) is reached at the last iteration. Note that, although the average number of iterations was rather small in this example, this is not usually the case with real datasets, where the underlying clustering structure is usually more complex (see Section \ref{sectionapplication}). 
	
	In sum, the iterative process described in Algorithm \ref{algorithm1} performs in a reasonable way, being able to discover the true clustering structure in barely 3 iterations in this example.

	\subsubsection{Employing the out-of-sample error in Algorithm \ref{algorithm1}}\label{eofsea1}
	
	In the numerical experiments carried out above, the in-sample error was employed to measure the distance from a series to a given cluster. This often works with global linear models, since the in-sample error is a reliable indicator of the predictive accuracy in this context. However, when more complex models are considered, the use of the in-sample error can lead to misleading results. For instance, a complex global model can reach zero in-sample error due to overfitting, and then generalize poorly over new observations. To avoid these undesirable situations, it is necessary to consider validation periods which are not used to fit the global models. 
	
	Based on previous comments, we decided to analyse the behaviour of Algorithm \ref{algorithm1} in the particular case of training and validation periods being disjoint. To this aim, we considered Scenario 2 but modified the training and validation sets. Specifically, for each series, training and validation periods were fixed to the first $(T-h-l_{\text{SIG}})$ observations and to observations from $(T-h-l_{\text{SIG}}+1)$ to $(T-h)$, respectively. In addition, the minimum series length was set to $T=100$ in this new setting due to the fact that $T=50$ would produce very short training periods. 
	
	Table \ref{tableoutofsampleari} contains the results of this new analysis in terms of ARI for method CPAGM. The performance of the method moderately decreased when considering the out-of-sample error. In fact, by comparing Table \ref{tableoutofsampleari} with Table \ref{tableari2}, it is clear that the clustering effectiveness of the proposed approach is better when the in-sample-error is employed, specially when only a few series per cluster are considered. The worse performance of CPAGM was expected, since fewer observations are used for both fitting the global models and performing the reassignation step. This decrease in sample size ends up causing a higher instability. In any case, the proposed method still outperforms the alternative approaches by a large degree in this new scenario. Results in terms of MAE are provided in Table \ref{tableoutofsamplemae} in the Appendix. In brief, CPAGM exhibits a worse behaviour for the shortest values of $T$, but the differences are minor with respect to the original Scenario 2.

	\begin{table}
		\centering
		\resizebox{3.5cm}{!}{\begin{tabular}{cc}  \hline 		
			$(T,N)$ & Average ARI  \\   \hline 
			$(100, 5)$  & 0.744   \\   
			$(100, 10)$  & 0.832   \\  
			$(100, 20)$  & 0.831   \\ 
			$(100, 50)$  & 0.876   \\  \hline
			$(200, 5)$  & 0.788  \\ 
			$(200, 10)$  & 0.841   \\  
			$(200, 20)$  & 0.859   \\  
			$(200, 50)$  & 0.871   \\  \hline
			$(400, 5)$  & 0.777   \\ 
			$(400, 10)$  & 0.810   \\ 
			$(400, 20)$ & 0.840  \\ 
			$(400, 50)$ & 0.877   \\   \hline
			$(1000, 5)$ & 0.785   \\  
			$(1000, 10)$ & 0.802  \\  
			$(1000, 20)$ & 0.845  \\  
			$(1000, 50)$ & 0.864   \\    \hline \\
		\end{tabular}}
		\caption{Average ARI Scenario 2 for method CPAGM. Out-of-sample error was used to assign the series to the clusters.}
		\label{tableoutofsampleari}
	\end{table}
	
	\subsubsection{Nonlinear global models}\label{subsubsectionnl}
	
	Throughout this section, the proposed clustering algorithm was assessed in scenarios where the different clusters are characterized by linear structures. The corresponding results indicated that, when the linearity assumption is met, the iterative procedure outlined in Algorithm \ref{algorithm1} shows an outstanding performance when linear global models fitted by least squares are considered. However, as real time series in many domains exhibit a certain degree of nonlinearity, it is interesting to evaluate the clustering technique in situations where the underlying stochastic processes are highly nonlinear. To this aim, an additional setup including the so-called self-exciting threshold autoregressive (SETAR) processes introduced by \cite{tong1980threshold} was considered. SETAR models can adequately describe many nonlinear features commonly observed in practice, as limit cycles and jump phenomena, among others. The specific generating structures in this new setting are provided below. \\
	
	\noindent \textbf{Scenario 3}. Let $\{X_t\}_{t \in \mathbb{Z}}$ be a stochastic process following the SETAR($p$)-type recursion given by
	
	\begin{equation}\label{setarp}
		X_t= \begin{cases}
			\beta_{0}^{(1)}+\sum_{i=1}^{p}\beta_{i}^{(1)}X_{t-i}+\epsilon_t^{(1)} & \text{if } X_{t-d} \le r, \\
				\beta_{0}^{(2)}+\sum_{i=1}^{p}\beta_{i}^{(2)}X_{t-i}+\epsilon_t^{(2)} & \text{if } X_{t-d} > r,
		\end{cases}
	\end{equation}

	\noindent where $\beta_0^{(j)}, \beta_1^{(j)}, \ldots,\beta_p^{(j)}$, $j=1,2$, are real numbers verifying the corresponding stationarity condition and $\{\epsilon_t^{(j)}\}_{t \in \mathbb{Z}}$, $j=1,2$, is a process formed by independent variables following the standard normal distribution. We fix $p=5$ and $d=3$. The vector of coefficients $\bm \beta_5=(\beta_0^{(1)}, \beta_1^{(1)}, \ldots,\beta_5^{(1)}, \beta_0^{(2)}, \beta_1^{(2)}, \ldots,\beta_5^{(2)})$ and the parameter $r$ are set as indicated below. \\

	\noindent Process 1: $\bm \beta_5=(0, 0.2, 0.9, -0.7, 0.3, -0.4, 0, 0.5, -0.6, 0.5, -0.4, 0.4)$, $r=1.2$. \\
	\noindent Process 2: $\bm \beta_5=(0, -0.2, -0.9, 0.7, -0.3, 0.4, 0, -0.5, 0.6, -0.5, 0.4, -0.4)$, $r=0$. \\
	\noindent Process 3: $\bm \beta_5=(0, 0.3, 0.3, 0.3, -0.4, -0.4, 0, -0.1, -0.7, -0.3, 0.5, 0.5)$, $r=0.6$. \\
	
	A new simulation experiment was designed to evaluate the proposed clustering algorithm in Scenario 3. As in previous analyses, several values of $N$ and $T$ were taken into account to generate the series. This time, due to the higher complexity of Scenario 3 in comparison with Scenarios 1 and 2, the test set was constructed by considering the last $h=l_\text{SIG}=5$ observations of each series. The out-of-sample error was employed as dissimilarity measure concerning the iterative procedure in Algorithm \ref{algorithm1}. Specifically, training and validation sets were defined as indicated in Section \ref{eofsea1}. The number of clusters was set to $K=3$. As Scenario 3 contains nonlinear processes, the random forest was selected for the global models involved in the approaches CPAGM, GMFBC and GMAP. The number of lags to fit the global models was set to $l=l_\text{SIG}=5$. With regards to the local approach (LM), a random forest was independently fitted to each one of the series. These individual models were used to compute the average prediction error. Note that the construction of a feature-based clustering approach based on the random forest is not straightforward. Therefore, the clustering accuracy associated with the LM method was obtained by considering the estimated coefficients of standard linear models. The simulation procedure was repeated 200 times and the ARI and the MAE were employed again as performance measures. 
	
	The results for Scenario 3 in terms of clustering effectiveness are provided in Table \ref{tableari3}. The value $N=50$ was not considered in this new simulation experiment due to the high computational cost of CPAGM when nonlinear global models as the random forest are fitted. The same statistical tests indicated in Section \ref{subsectionresultsdiscussion} were carried out along with the corresponding Bonferroni corrections. According to Table \ref{tableari3}, the proposed algorithm attains significantly higher ARI values than the alternative ones in most cases, with a clustering accuracy which generally increases with the series length ($T$). This effect is not observed for the number of series per process ($N$), which is probably due to the complexity of the models in this new scenario. The approach GMFBC shows a rather poor performance, which suggests that the features employed by this method are not appropriate to detect the dependence structure of the SETAR processes in Scenario 3. The LM approach also exhibits low scores, which was expected, since it considers estimated features based on linear models.

	\begin{table}
		\centering
		\resizebox{6cm}{!}{\begin{tabular}{cccc}  \hline 		
				$(T,N)$ & LM & CPAGM & GMFBC \\   \hline 
				$(50, 5)$  & 0.250 & \textbf{0.321}$^*$ & 0.130 \\   
				$(50, 10)$  & 0.231 & \textbf{0.357}$^*$ & 0.109 \\  
				$(50, 20)$  & 0.149 & \textbf{0.354}$^*$ & 0.054 \\   \hline
				$(100, 5)$  & 0.278 & \textbf{0.511}$^*$ & 0.198 \\   
				$(100, 10)$  & 0.259 & \textbf{0.531}$^*$ & 0.128 \\  
				$(100, 20)$  & 0.201 & \textbf{0.540}$^*$ & 0.080 \\  \hline
				$(200, 5)$  & 0.267 & \textbf{0.543}$^*$ & 0.181 \\ 
				$(200, 10)$  & 0.225 &\textbf{0.582}$^*$ & 0.143 \\  
				$(200, 20)$  & 0.220 & \textbf{0.535}$^*$ & 0.060 \\  \hline
				$(400, 5)$  & 0.347 & \textbf{0.708}$^*$ & 0.297 \\ 
				$(400, 10)$  & 0.290 & \textbf{0.675}$^*$ & 0.207 \\ 
				$(400, 20)$ & 0.148 & \textbf{0.638}$^*$ & 0.059 \\  \hline
				$(1000, 5)$ & 0.397 & \textbf{0.741}$^*$ & 0.312 \\  
				$(1000, 10)$ & 0.245 & \textbf{0.732}$^*$ & 0.145 \\  
				$(1000, 20)$ & 0.231 & \textbf{0.738}$^*$ & 0.135 \\   \hline 
		\end{tabular}}
		\caption{Average ARI in Scenario 3. For each pair $(T, N)$, the best result is shown in bold. An asterisk indicates that a given method is significantly better than the rest at level $\alpha=0.01$.}
		\label{tableari3}
	\end{table}

Results in terms of predictive accuracy are given in Table \ref{tablemae3} in the Appendix. The corresponding values indicate that CPAGM exhibits a significantly lower forecasting error than the remaining approaches in most of the settings.

In sum, the analyses carried out throughout this section illustrate the flexibility of the proposed clustering approach, which can deal with series generated from highly complex processes as long the class of global models is chosen appropriately. 
	
	\section{Application to real data}\label{sectionapplication}
	
	In this section, we apply the proposed algorithm to perform clustering in several well-known datasets. All of them have been used in many peer-reviewed publications as standard benchmarks, from literature on local models to recent works on global models. Specifically, \cite{montero2021principles} employed these databases to show the advantages of global methods over local methods in terms of predictive accuracy. After describing the data, the method is first applied to each one of the databases individually. Then, we show an application in which series from two databases are combined. It is worth highlighting that, for the sake of simplicity and computational efficiency, the analyses shown throughout this section are focused on global linear models. 
	
	\subsection{Applying the proposed algorithm to each one of the datasets independently}\label{subsectionapplication1}
	
	This section shows the application of the proposed approach to some real time series databases which pertain in turn to some data collections described below. 
	
	\begin{itemize}
		\item \textbf{M1}. Heterogeneous dataset from a forecasting competition \cite{makridakis1982accuracy}. It contains 1001 series subdivided in yearly (181), quarterly (203) and monthly (617) periodicity. The considered datasets are referred to as M1 Yearly, M1 Quarterly and M1 Monthly, respectively. 
		\item \textbf{M3}. Heterogeneous database from a forecasting competition \cite{makridakis2000m3}, containing 3003 time series subdivided in yearly (645), quarterly (756), monthly (1428) and an extra category of periodicity, so-called ``other'' (174).  The considered datasets are referred to as M3 Yearly, M3 Quarterly, M3 Monthly and M3 Other, respectively. 
		\item \textbf{Tourism}. Homogeneous dataset from a tourism forecasting competition \cite{athanasopoulos2011tourism}, including 1311 series divided into yearly (518), quarterly (427) and monthly (366) data. The considered datasets are referred to as Tourism Yearly, Tourism Quarterly and Tourism Monthly, respectively. 
	\end{itemize}
	
	Method CPAGM and the alternative approaches examined in Section \ref{sectionsimulationstudy} were executed in each one of the previous datasets. No data preprocessing was performed, since there is not a clear agreement about the benefits of preprocessing when fitting global models \cite{montero2021principles}. It is worth noting that, unlike in the simulation study, there is no way of objectively assessing the quality of the clustering partition in these databases, since no information about the ground truth is available. Hence, our comparative analyses focus on the predictive effectiveness of the considered techniques. In all cases, the test sets were constructed by considering the last $h = 5$ observations of each time series. Procedures CPAGM, GMFBC and GMAP were run for several values of $K$, namely $K \in \{1, 2, 3, 4, 5, 7, 10\}$ and $l$. Note that the range of the latter parameter is limited by the minimum series length existing in a given database.
	
	To measure the predictive accuracy, we considered two well-known error metrics, namely the mean absolute scaled error (MASE) and the symmetric mean absolute percentage error (sMAPE). Using a scale-free or a percentage error is desirable in our setting because, unlike in the numerical experiments of Section \ref{sectionsimulationstudy}, some databases contain series which are recorded in very different scales. Thus, employing the MAE could have resulted in the average forecasting error being corrupted by the higher influence of the series in the largest scales. Note that, by considering the MASE and sMAPE metrics, the average prediction error in \eqref{errormetric} takes the form
	
	\begin{equation}\label{errormetricsmape}
		\begin{split}
			\frac{1}{n}\sum_{k=1}^{K}\sum_{{\substack{i=1:\\ \bm X_t^{(i)} \in C_k}}}^{n}d^*_{\text{MASE}}\big(\bm{X}_t^{(i)*}, \overline{\mathcal{M}}_k\big)  \, \, \, \text{and} \, \, \,  \frac{1}{n}\sum_{k=1}^{K}\sum_{{\substack{i=1:\\ \bm X_t^{(i)} \in C_k}}}^{n}d^*_{\text{sMAPE}}\big(\bm{X}_t^{(i)*}, \overline{\mathcal{M}}_k\big),
		\end{split}
	\end{equation}

\noindent respectively, where

\begin{equation}\label{errortestset}
\begin{split}
	d^*_{\text{MASE}}\big(\bm{X}_t^{(i)*}, \overline{\mathcal{M}}_k\big)=\frac{\frac{1}{h}\sum_{j=1}^{h}\big|X_j^{(i)*}-\overline{F}_{j,k}^{(i)*}\big|}{\text{MAE}^i_{\text{Naive}}},\\
d^*_{\text{sMAPE}}\big(\bm{X}_t^{(i)*}, \overline{\mathcal{M}}_k\big)=\frac{200}{h}\sum_{j=1}^{h}\Bigg(\frac{\big|X_j^{(i)*}-\overline{F}_{j,k}^{(i)*}\big|}{\big|X_j^{(i)*}\big|+\big|\overline{F}_{j,k}^{(i)*}\big|}\Bigg),
\end{split}
\end{equation}

\noindent for $i=1,\ldots,n$, $k=1, \ldots, K$, with $\text{MAE}^i_{\text{Naive}}=\frac{1}{L_i-m_i}\sum_{t=m_i+1}^{L_i}\big|X_t^{(i)}-X_{t-m_i}^{(i)}\big|$ and $m_i$ denoting the seasonal period of the $i$th time series ($m_i=1$ if the series is nonseasonal). 

\begin{figure}
		\centering
		\includegraphics[width=1\textwidth]{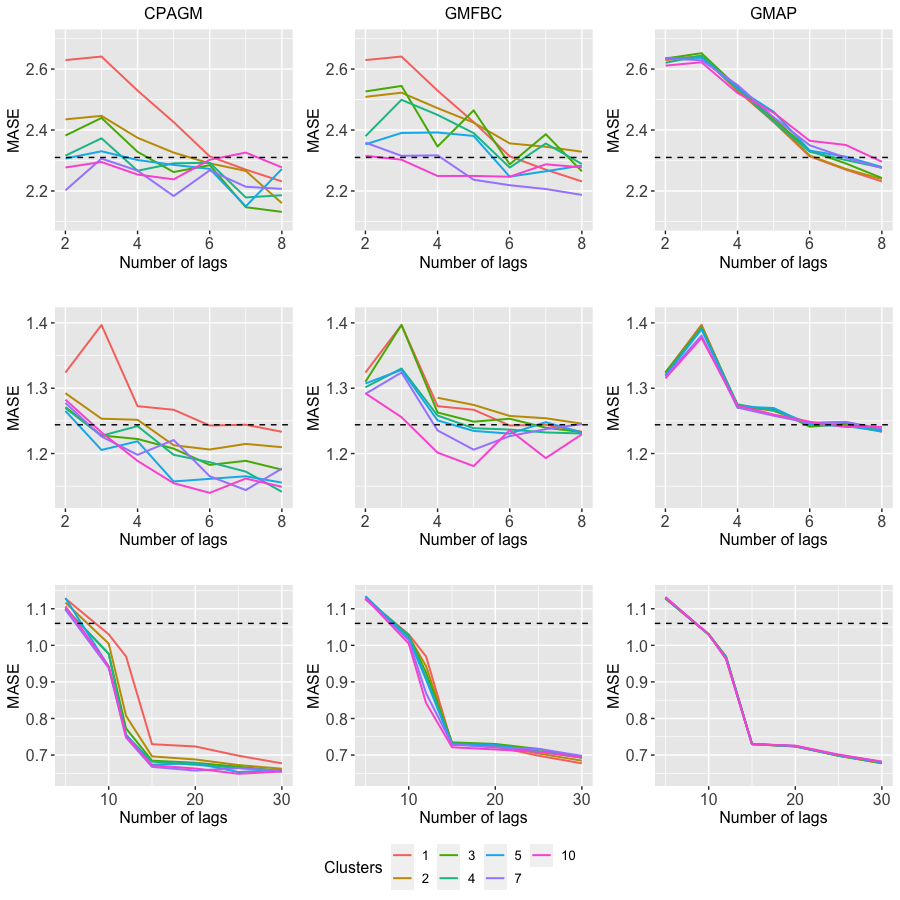}
		\caption{Average MASE as a function of the number of lags in datasets M1 Yearly (top panel), M1 Quarterly (middle panel) and M1 Monthly (bottom panel). Each color corresponds to a different value for the number of clusters, $K$. The horizontal dashed line indicates the average MASE achieved by the approach based on local models (LM).}
		\label{m1plots}
	\end{figure}

\begin{figure}
	\centering
	\includegraphics[width=1\textwidth]{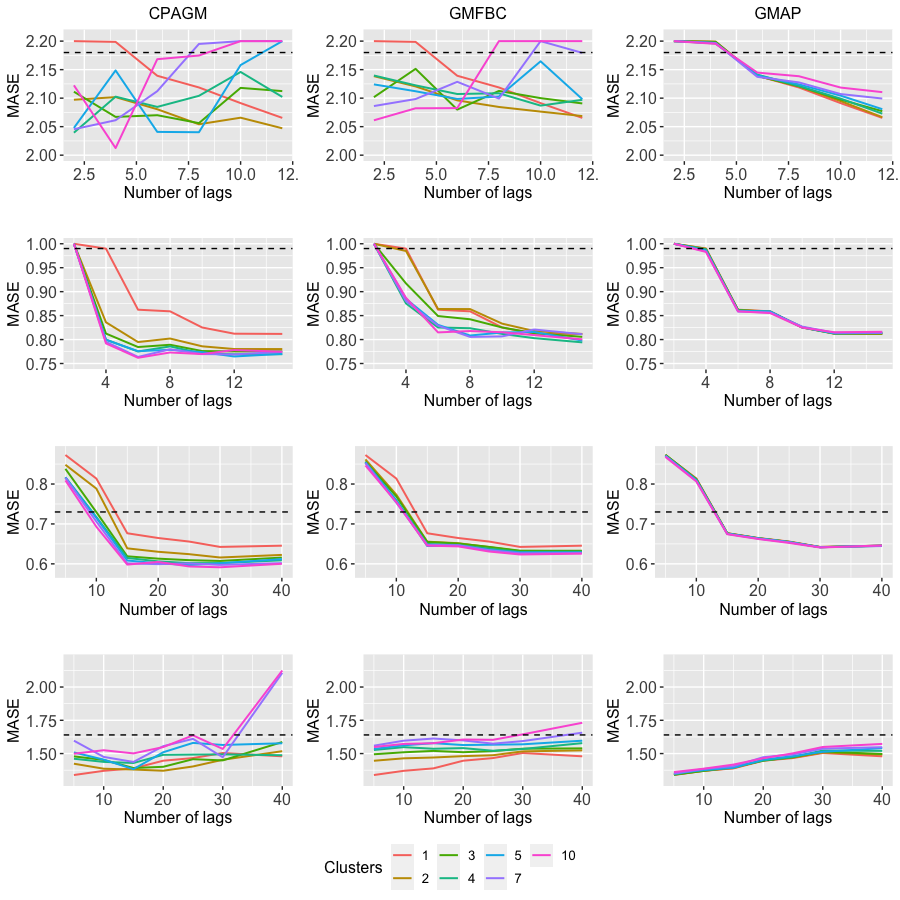}
	\caption{Average MASE as a function of the number of lags in datasets M3 Yearly (top panel), M3 Quarterly (upper middle panel), M3 Monthly (lower middle panel) and M3 Other (bottom panel). Each color corresponds to a different value for the number of clusters, $K$. The horizontal dashed line indicates the average MASE achieved by the approach based on local models (LM).}
	\label{m3plots}
\end{figure}

\begin{figure}
	\centering
	\includegraphics[width=1\textwidth]{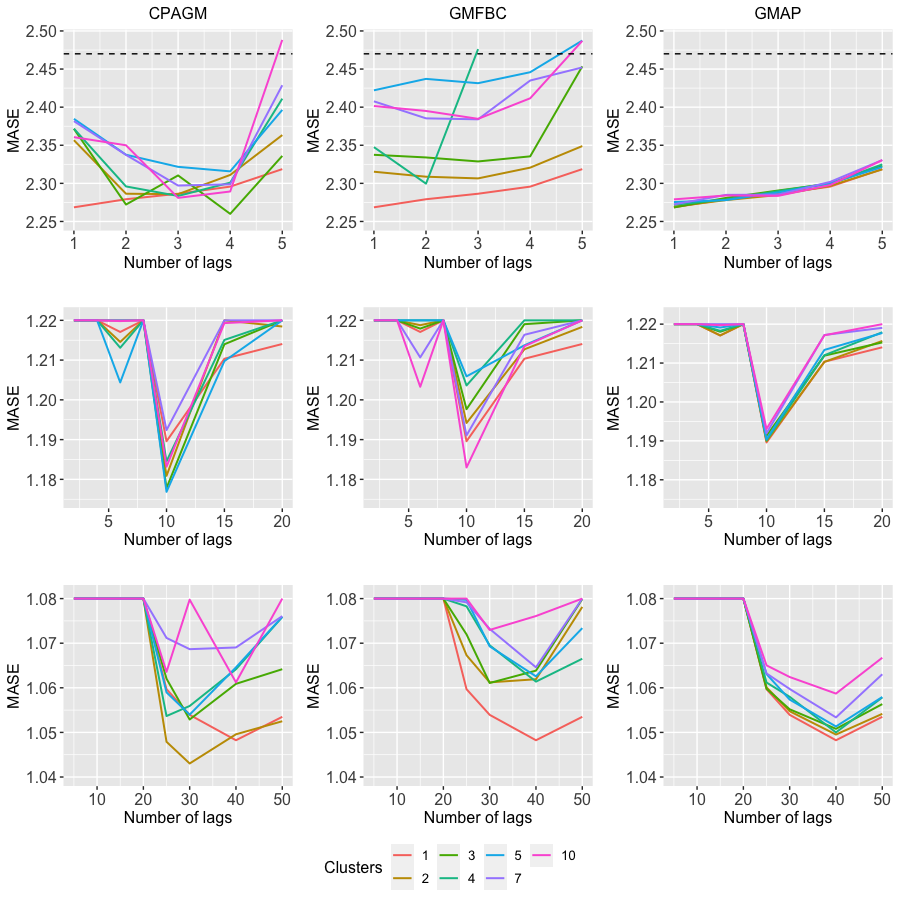}
	\caption{Average MASE as a function of the number of lags in datasets Tourism Yearly (top panel), Tourism Quarterly (middle panel) and Tourism Monthly (bottom panel). Each color corresponds to a different value for the number of clusters, $K$. The horizontal dashed line indicates the average MASE achieved by the approach based on local models (LM).}
	\label{tourismplots}
\end{figure}

Figures \ref{m1plots}, \ref{m3plots} and \ref{tourismplots} show the results in terms of MASE for datasets pertaining to M1, M3 and Tourism collections, respectively. For a given method, curves of average MASE were represented as a function of the number of lags. Several colors were used to indicate the different values of $K$. The average forecasting error associated with the LM approach was incorporated to a given graph by means of an horizontal dashed line. In the plots associated with the databases Tourism Quarterly and Tourism Monthly (middle and bottom panels in Figure \ref{tourismplots}), these horizontal lines are not visible because the local approach exhibits a rather high forecasting error.      
	
The graphs in Figures \ref{m1plots} and \ref{m3plots} indicate that, in datasets belonging to M1 and M3 collections, splitting the series into different clusters by means of the proposed approach is advantageous. In fact, the red curve, which corresponds to $K=1$, is usually above all the remaining curves (the only exception being dataset M3 Other). This indicates that a better prediction accuracy is achieved when the series are grouped according to the underlying forecasting structures. In addition, a general pattern of reduction in the forecasting error is usually observed when increasing the order of the global models. Note that the proposed algorithm substantially outperforms the approach based on local models for several values of the number of clusters and the number of lags, sometimes over the whole range (e.g., dataset M3 Quarterly), which is consistent with the conclusions of \cite{montero2021principles} concerning the effectiveness of global models in the considered datasets. Splitting the series into different groups according to the approach of \cite{bandara2020forecasting} is also generally advantageous, thus indicating that the features employed by this method contain useful information about the forecasting structures of the time series. However, the corresponding average MASE is usually greater than the one produced by CPAGM, since the latter method is specifically designed to generate an optimal partition in terms of predictive accuracy. As expected, the approach based on an arbitrary partition (GMAP) shows a worse behavior. As this method splits the series totally at random, its forecasting error is usually similar to the one produced by a single global model ($K=1$). In fact, in some datasets (e.g., M3 Yearly), the performance of this approach clearly worsens when increasing the number of groups, which is reasonable, since the consideration of a larger number of random clusters means a worse exploitation of the information contained in the database.  

A different situation happens for datasets pertaining to Tourism collection (Figure \ref{tourismplots}). In fact, in two of these databases, namely Tourism Yearly and Tourism Monthly, grouping the series into different clusters does not result in a better predictive accuracy. Note that this is not an issue, since one can not expect the proposed method to be advantageous in each and every database. It is worth remarking that, in these datasets, approaches CPAGM and GMFBC exhibit a higher prediction error than the method based on a random partition (GMAP) when $K>1$. This means that, in such cases, partitioning the set according to a given criterion aimed at maximizing the predictive accuracy is counterproductive. This could be due to the fact that, in most of the corresponding time series, the observations constituting the test periods behave rather differently from the remaining ones. In any case, an in-depth analysis of the series in these data collections would be desirable. On the other hand, method CPAGM outperforms GMFBC and GMAP in dataset Tourism Quarterly. Specifically, a number of lags of $l=10$ is the optimal one for this approach regardless of the number of clusters. In fact, values of $l$ above or below this value result in a higher forecasting error. 

In order to better understand the results of Figures \ref{m1plots}, \ref{m3plots} and \ref{tourismplots}, average values of MASE are provided in Table \ref{errordatasets}. Specifically, for approaches CPAGM, GMFBC and GMAP, we report the forecasting error associated with the optimal pair $(K,l)$. The corresponding quantities indicate that the proposed method exhibits a significant advantage over the remaining approaches in most datasets of M1 and M3 collections (with the only exception of M3 Other) but results only in small improvements in datasets pertaining to Tourism collection. Results in terms of sMAPE error were also incorporated to Table \ref{errordatasets}. The superiority of the proposed approach over the alternative methods is generally more pronounced according to this new error metric. In fact, in some datasets (e.g., M1 Monthly), the differences with respect to the feature-based approach GMFBC are dramatic. Interestingly, CPAGM clearly achieves the lowest average sMAPE in two datasets pertaining to Tourism collection, namely Tourism Quarterly and Tourism Monthly. In addition, there are some datasets in which the proposed method does not show the best performance. For instance, in Tourism Yearly, the local approach exhibits by far the highest predictive accuracy. Finally, it is worth highlighting that similar conclusions to the ones stated above were obtained when considering longer test periods ($h>5$).  

\begin{table}
	\centering
		\resizebox{10cm}{!}{\begin{tabular}{cccccc}\hline 
Dataset	&	Measure    & LM & CPAGM ($K=1$) & GMFBC & GMAP \\ \hline 
	M1 Yearly &	MASE &    2.310 & \textbf{2.131} (2.231) &  2.187     &   2.231   \\
   &	sMAPE &     138.550 & 75.574 (100.683) &  \textbf{71.829 }    &   92.362   \\ \hline 
	M1 Quarterly &	MASE &  1.244  &   \textbf{1.139} (1.233)         &   1.181    &  1.233     \\
	 &	sMAPE &  26.266  &   \textbf{21.645} (75.518)         &   44.467    &  60.842     \\ \hline 
	M1 Monthly &	MASE &    1.060 & \textbf{0.649} (0.677) &  0.677     &   0.677   \\
	&	sMAPE &    18.511 & \textbf{16.609} (64.778) &  42.946     &   52.703   \\ \hline 
	M3 Yearly &	MASE &  2.182  &     \textbf{2.012} (2.065)        &   2.061    &  2.065    \\
	 &	sMAPE &  15.546  &   15.598 (15.874)         &   \textbf{15.384}    &  15.874    \\ \hline 
	M3 Quarterly &	MASE &  0.992  &      \textbf{0.761} (0.812)       & 0.794       & 0.812     \\
	 &	sMAPE &  8.460  &      \textbf{7.069} (7.965)       & 7.440       & 7.965     \\ \hline 
	M3 Monthly &	MASE & 0.732   & \textbf{0.591} (0.642) &   0.623    &  0.641    \\
	 &	sMAPE & 13.141 & \textbf{11.918} (12.467) &   12.256    &  12.467    \\ \hline 
	M3 Other &	MASE &  1.639  &     \textbf{1.339} (\textbf{1.339})        &   \textbf{1.339}    & \textbf{1.339}     \\
	 &	sMAPE &  4.046  &     \textbf{3.215} (3.729)        &   3.657    & 3.729     \\ \hline 
	Tourism Yearly &	MASE &  2.470  &     \textbf{2.260} (2.269)        &  2.269     &  2.268    \\  
	 &	sMAPE &  \textbf{22.434}  &     39.956 (39.956)        &  39.956     &  39.956    \\   \hline 
	Tourism Quarterly &	MASE &  2.501  & \textbf{1.177} (1.190) & 1.183 & 1.190     \\
	 &	sMAPE &  22.101  & \textbf{14.515} (19.571) & 19.571 & 19.571     \\ \hline 
	Tourism Monthly &	MASE &  2.327  & \textbf{1.043} (1.048) &   1.048    &  1.048 \\ 
	 &	sMAPE &  30.349  & \textbf{16.305} (18.494) &   18.494    &  18.494    \\ \hline  
	\end{tabular}}
\caption{Average MASE and sMAPE associated with the optimal pair $(K, l)$ for methods CPAGM, GMFBC and GMAP. The average errors obtained by the LM approach were also incorporated. For each dataset and error metric, the best result is shown in bold}
\label{errordatasets}
\end{table}

\begin{figure}
	\centering
	\includegraphics[width=0.8\textwidth]{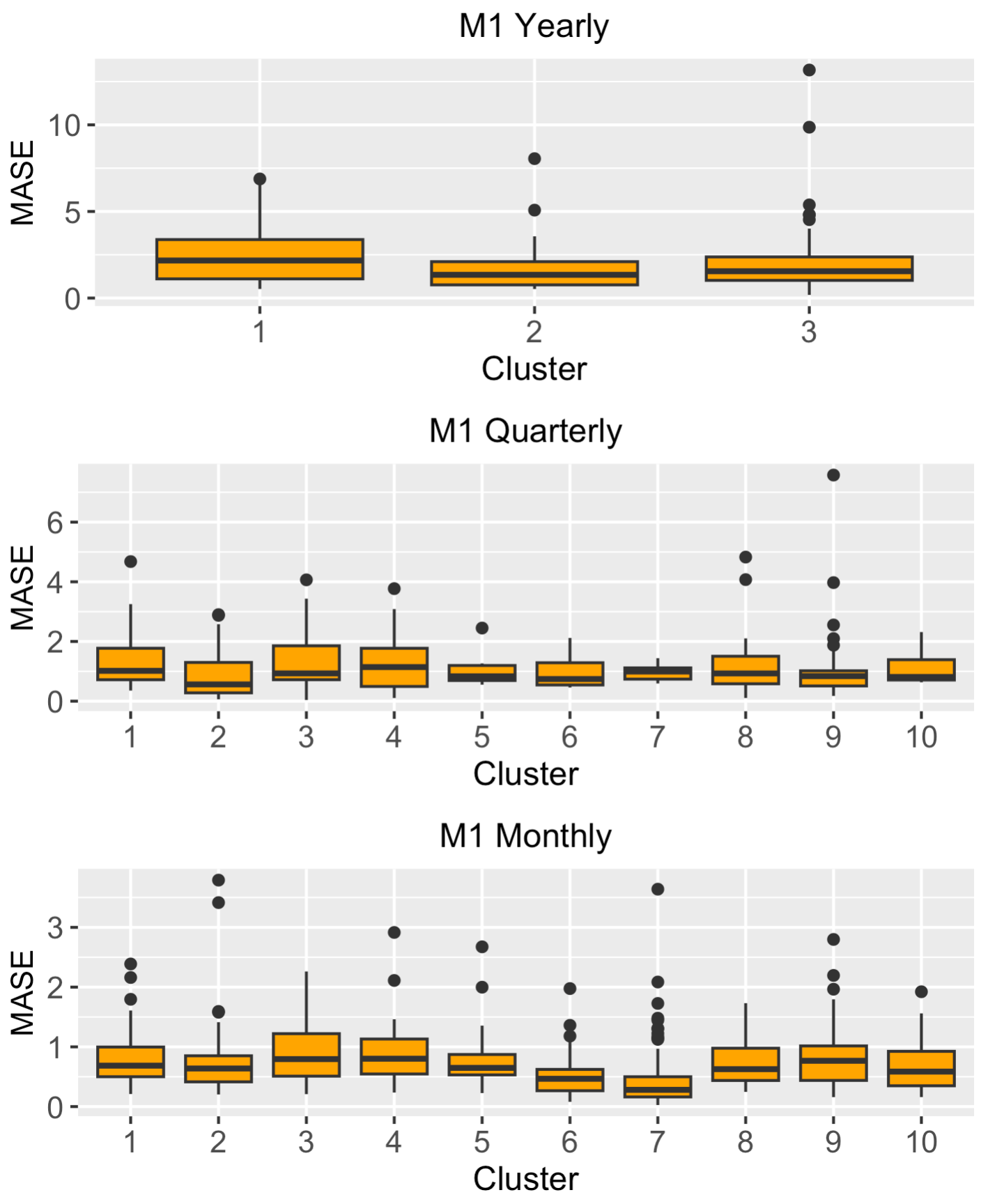}
	\caption{Distribution of MASE for the different clusters concerning the partitions produced by the proposed method in datasets M1 Yearly (top panel), M1 Quarterly (middle panel) and M1 Monthly (bottom panel). The optimal values for $K$ and $l$ were considered for each dataset.}
	\label{boxplotsm1}
\end{figure}

As shown in Section \ref{sectionclusteringalgorithm}, the proposed method constructs a clustering partition in a way that the overall prediction error is minimized. This quantity is computed by obtaining the forecasts of each time series using the global model (prototype) associated with its group, which are then compared with the corresponding test periods (see \eqref{errormetric}). Thus, each global model has a certain contribution to the overall prediction error in the form of individual error terms, and a way of assessing the quality of the model consists of examining the distribution of these quantities. To this aim, a boxplot can be constructed for each one of the groups by using the final prediction errors produced by the clustering algorithm. The top, middle and bottom panels of Figure \ref{boxplotsm1} provide the corresponding boxplots for datasets M1 Yearly, M1 Quarterly and M1 Monthly, respectively, which were constructed by considering the optimal values of $K$ and $l$, and the MASE as the error metric. In the three cases, the distribution of the forecasting error is clearly different among groups. For instance, in dataset M1 Yearly, the series in the second and third clusters usually yield better predictions than the series in the first group. Moreover, the corresponding prediction errors show more variability in the latter case. These properties suggest a higher degree of similarity for the series in the second and third clusters in terms of forecasting structures, which results in more accurate global models. Similar conclusions can be reached by analyzing the boxplots for datasets M1 Quarterly and M1 Monthly. Note that, in the latter database, there are several time series giving rise to extremely large values of the forecasting error, which could require an individual analysis.  

A numerical summary of the boxplots in Figure \ref{boxplotsm1} is given in Table \ref{tablem1}. In particular, for each cluster, the relative size is provided along with the sample mean and variance of the corresponding MASE values. These quantities corroborate the existence of substantial differences among the groups in a given dataset. Note that, in datasets M1 Yearly and M1 Monthly, the cluster containing the largest number of series is associated with the lowest average forecasting error. This is reasonable, since more series imply more accurate estimates for the coefficients of the underlying global model, thus resulting in a better predictive accuracy. 

\begin{table}
	\centering 
	\resizebox{11cm}{!}{\begin{tabular}{ccccccccccc}\hline 
		M1 Yearly   &   &   &   &  &  &  &  &  &  &  \\
		Cluster         & 1 & 2 & 3 &  - & - & - & - & - & - & - \\ \hline  
		Relative size   & 0.38  & 0.40  & 0.22  &  - & - & - & - & - & - & -  \\
		Mean (MASE)     & 1.14  &  0.83  & 1.16  &  - & - & - & - & - & - & -  \\
		Variance (MASE) &  0.59 &  0.55 & 0.52  &  - & - & - & - & - & - & -   \\ \hline 
		M1 Quarterly   &   &   &   &  &  &  &  &  &  &  \\
		Cluster         & 1 & 2 & 3 &  4 & 5 & 6 & 7 & 8 & 9 & 10 \\ \hline  
		Relative size   &  0.06 & 0.15  &  0.15 & 0.13 & 0.06 &  0.06 & 0.02 & 0.10  & 0.22 & 0.04  \\
		Mean (MASE)     &  1.52 & 0.91  & 1.33  & 1.32 & 1.03 & 0.97 & 0.97 & 1.31 & 1.06 & 1.13  \\
		Variance (MASE) & 1.66  & 0.73  & 1.15  & 0.96 & 0.25 & 0.32 & 0.11 & 0.35 & 0.48 & 0.34  \\ \hline 
		M1 Monthly   &   &   &   &  &  &  &  &  &  &  \\
		Cluster         & 1 & 2 & 3 &  4 & 5 & 6 & 7 & 8 & 9 & 10 \\ \hline  
		Relative size   &   0.07 & 0.13   & 0.06  & 0.05  & 0.05 & 0.15  & 0.26 & 0.07 & 0.07 & 0.09  \\
		Mean (MASE)     & 0.85  & 0.74  & 0.92  & 0.89 & 0.79 & 0.50 & 0.41 & 0.73 & 0.87 & 0.69 \\
		Variance (MASE) &  0.25 &  0.32 &  0.30 & 0.33 & 0.27 & 0.10  & 0.19 & 0.13  & 0.33 & 0.17  \\ \hline 
	\end{tabular}}
\caption{Description of the different clusters concerning the partitions produced by the proposed method in datasets M1 Yearly (upper part), M1 Quarterly (middle part) and M1 Monthly (lower part). The optimal values for $K$ and $l$ were considered for each dataset.}
\label{tablem1}
\end{table}

\begin{figure}
	\centering
	\includegraphics[width=0.8\textwidth]{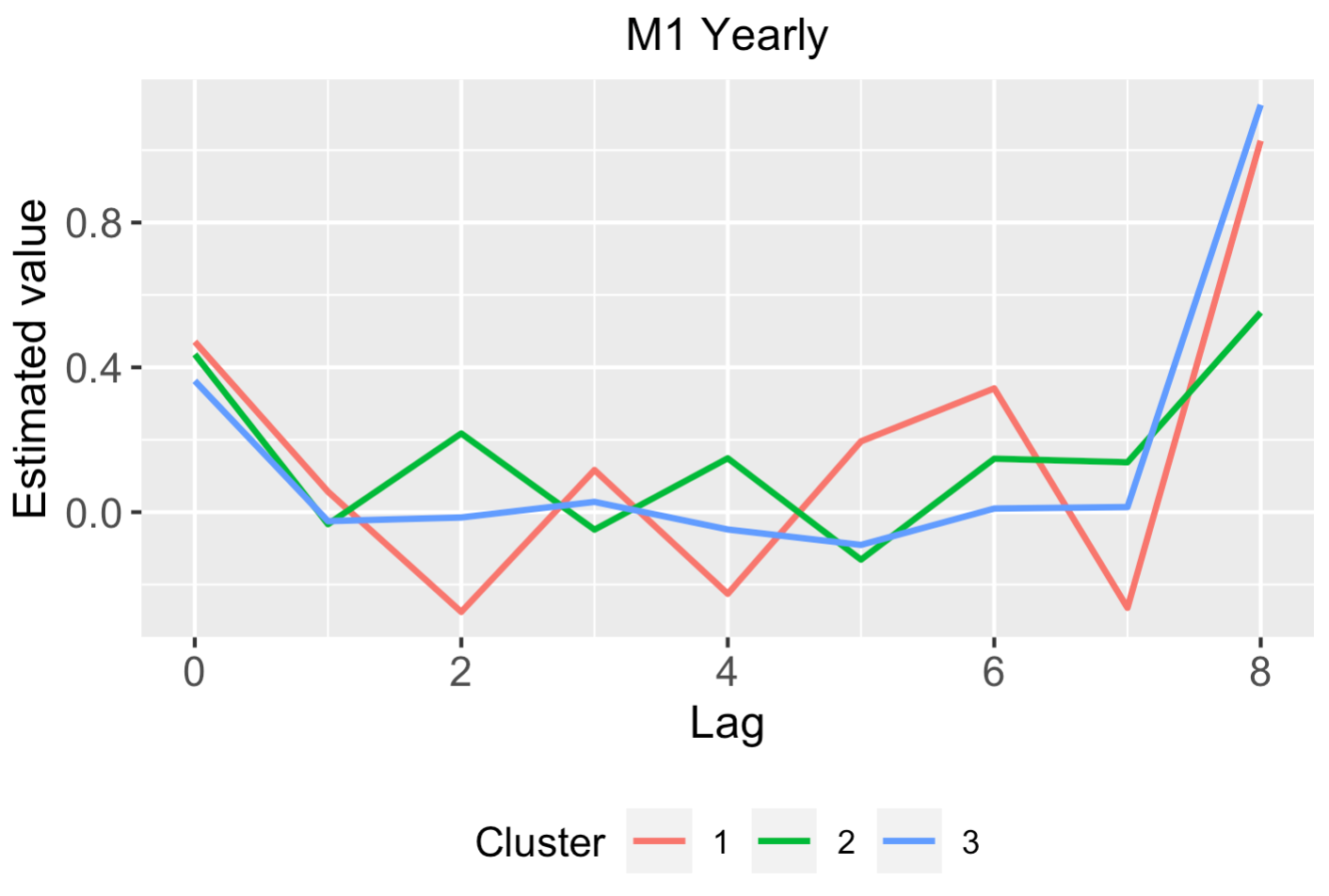}
	\caption{Estimated coefficients for lags 0 (intercepts) to 8 for the global linear models concerning the 3-cluster solution produced by CPAGM ($l=8$) in dataset M1 Yearly.}
	\label{lagsm1}
\end{figure}

Note that, besides the clustering partition, an essential element of the proposed clustering algorithm are the resulting prototypes, i.e., the final global models. In fact, these models characterize the forecasting structures of the different clusters, and their analysis can provide a meaningful description of the time series belonging to each group. Based on previous comments, we decided to examine the prototypes for the 3-cluster solution in dataset M1 Yearly (top panel of Figure \ref{boxplotsm1}). In this regard, Figure \ref{lagsm1} represents the estimated coefficients for the prototypes of the three clusters. Note that a number of 8 lags were considered to fit the global models. The values associated with $l=0$ in Figure \ref{lagsm1} indicate the estimates for the corresponding intercepts. Note that the three prototypes exhibit clearly dissimilar behaviors. In fact, the estimated coefficients for the global models of first and second clusters take rather different values for several lags, usually showing opposite signs. On the contrary, the global model of the third group lies somewhere in the middle, with estimated coefficients close to zero for lags between 1 and 7. In fact, the only significant lag for this prototype seems to be $l=8$, since the corresponding estimate takes a large value. Hence, the third cluster is expected to contain mostly series exhibiting significant serial dependence only at lag 8. Note that this is an important insight that can lead to interesting conclusions about the series in this group. In sum, the graph in Figure \ref{lagsm1} provides a useful decomposition of the linear forecasting structures existing in dataset M1 Yearly. 

\subsection{Additional analysis. Combining series from two datasets}\label{subsectionapplication2}

In the previous section, the proposed method and the alternative approaches were applied by considering each database independently. However, it is interesting to assess the performance of the different techniques with datasets containing series from different domains, since several real time series databases have this property. To this aim, we employed the data collections in Section \ref{subsectionapplication1} and created new databases by combining two of the former data collections. Algorithm CPAGM and the alternative approaches were executed with these two data collections. The results of this analysis are provided in the Appendix.

\section{Conclusions and future work}\label{sectionconclusions}
	
	In this work, a clustering algorithm based on prediction accuracy of global forecasting models was introduced. The procedure is based on an iterative mechanism and relies on the following two steps:
	
	\begin{enumerate}
		\item $K$ global models (prototypes) are fitted by considering the series belonging to each cluster.
		\item Each time series is assigned to the group associated with the prototype yielding the lowest forecasting error according to the MAE metric. 
	\end{enumerate} 
	
	Since the algorithm is specifically designed to minimize the overall prediction error, the resulting partition distributes the time series in such a way that the corresponding global models represent in the best possible way the existing forecasting structures. The method is motivated by the fact that, given two different model-based clustering solutions, the one generating the best predictions is preferred. In short, our method produces a meaningful clustering partition while providing a powerful tool to predict future values of the series. It is important to emphasize that, to assess the predictive ability of the procedure, a test period must be considered for each of the series in the collection. Otherwise, the forecasting error is likely to be underestimated. 
	
	Although several distance measures have been proposed in the literature to perform TSC (metrics based on geometric characteristics, extracted features, estimated model coefficients etc), to the best of our knowledge, no previous works have proposed to measure dissimilarity as the forecasting error produced by a global model. Moreover, the concept of prototype introduced in this manuscript, namely a global model fitted to all the time series within a given cluster, is also novel. In short, our method takes advantage of the outstanding performance of global models to find groups of series sharing the same forecasting patterns, a situation that can easily happen in real databases. It is important to remark that, although an improvement in the overall predictive accuracy is frequently attained by means of the proposed approach, the main output of the procedure is the resulting clustering solution, which produces a meaningful decomposition of the collection of time series in terms of forecasting structures and can be very useful as a exploratory tool. 
	
	The proposed approach was evaluated by means of a broad simulation study where the groups were characterized by different underlying stochastic processes. Some alternative methods as one procedure based on local models were considered for comparison purposes. The algorithm was also applied to perform clustering in classical time series datasets. Several important elements were analysed, including the number of clusters, the AR order of the global models, the class of models (linear regression, random forest etc) or the number of iterations the procedure needs to reach convergence. Overall, the proposed technique showed an excellent behaviour in terms of both clustering accuracy and forecasting effectiveness, outperforming the alternative approaches. 
	
	It is worth highlighting that the proposed procedure has some limitations. An important one is that, as the method considers the future periods of the series to compute the distance between each element and each prototype, the value of the objective function does not necessarily decrease with each iteration. However, this issue can be easily solved by means of a simple heuristic rule. In addition, there are some situations in which our method is not advantageous. For instance, if the considered dataset contains long time series exhibiting simple dependence structures, then an approach based on local models can get similar or even better results. Moreover, although the proposed algorithm gets great results in the simulation experiments with well-separated groups, its clustering accuracy decreases when some amount of uncertainty exists in the generating processes. Another important issue is related to the class of global models, which must be appropriately chosen for a correct identification of the underlying groups.  
	
	There are two main ways through which this work can be extended. First, the numerical issues related to the two-step iterative process in Algorithm \ref{algorithm1} could be addressed. In this regard, a new objective function could be proposed so that the forecasting error automatically decreases with each iteration, as in the traditional iterative clustering approaches. Second, our approach could be extended to the fuzzy setting. This way, each series would belong simultaneously to all the clusters and the corresponding forecasts would be computed as a weighted average by considering the individual predictions produced by each global model. This would probably result in increased stability of the overall forecasting error. Both topics will be properly addressed in further research. 
	
	\bibliography{mybibfile.bib}

 \clearpage 
	
\section*{Appendix}

\subsection*{Results in terms of predictive accuracy for the simulation experiments of Section \ref{sectionsimulationstudy}}

Average values of MAE associated with the different techniques in Scenario 1 are provided in Table \ref{tablemae1}. In order to perform rigorous comparisons, pairwise paired $t$-tests were carried out by taking into account the 200 simulation trials. In all cases, the alternative hypotheses stated that the mean MAE value of a given method is less than the mean MAE value of its counterpart. Bonferroni corrections were applied to the set of $p$-values associated with each value of $T$. An asterisk was incorporated in Table  \ref{tablemae1} if the corresponding method resulted significantly more effective than the remaining ones for a significance level 0.01. The results associated with running the approach CPAGM with $K=1$ (only one global model) were incorporated to Table \ref{tablemae1} by indicating ``($K=1$)''.

\begin{table}
	\centering
	\resizebox{8cm}{!}{\begin{tabular}{ccccc}  
			\hline $(T,N)$ & LM & CPAGM ($K=1$) &  GMFBC & GMAP  \\   \hline 
			(20, 5) & 1.066 & \textbf{1.043}  (1.069) & 1.072 & 1.078 \\
			(20, 10) & 1.068 & \textbf{0.997}$^*$  (1.075) & 1.046 & 1.080 \\   
			(20, 20) & 1.070 & \textbf{0.964}$^*$ (1.076) & 1.036 & 1.052 \\   
			(20, 50) & 1.073 & \textbf{0.942}$^*$  (1.075) & 1.034 & 1.046 \\ \hline
			(50, 5) & 1.019 & \textbf{0.921}$^*$  (1.065) & 1.011 & 1.100 \\   
			(50, 10) & 1.023 & \textbf{0.913}$^*$  (1.073) & 1.021 & 1.044 \\   
			(50, 20) & 1.024 & \textbf{0.910}$^*$  (1.082) & 1.024 & 1.072 \\   
			(50, 50) & 1.016 & \textbf{0.907}$^*$  (1.074) & 1.020 & 1.042 \\  \hline  
			(100, 5) & 0.976 & \textbf{0.919}$^*$  (1.072) & 0.994 & 1.225 \\  
			(100, 10) & 0.978 & \textbf{0.913}$^*$  (1.075) & 0.996 & 1.148 \\  
			(100, 20) & 0.976 & \textbf{0.911}$^*$  (1.076) & 1.003 & 1.067 \\   
			(100, 50) & 0.977 & \textbf{0.911}$^*$  (1.079) & 1.009 & 1.061 \\ \hline 
			(200, 5) & 0.929 & \textbf{0.911} (1.062) & 0.949 & 1.025 \\   
			(200, 10) & 0.942 & \textbf{0.918}$^*$  (1.083) & 0.968 & 1.058 \\   
			(200, 20) & 0.938 & \textbf{0.912}$^*$  (1.070) & 0.969 & 1.062 \\   
			(200, 50) & 0.942 & \textbf{0.916}$^*$  (1.073) & 0.978 & 1.090 \\ \hline 
			(400, 5) & 0.920 & \textbf{0.915}  (1.069) & 0.937 & 1.092 \\   
			(400, 10) & 0.920 & \textbf{0.916}  (1.076) & 0.937 & 1.069 \\   
			(400, 20) & 0.929 & \textbf{0.926}  (1.080) & 0.949 & 1.071 \\   
			(400, 50) & \textbf{0.925} & \textbf{0.925} (1.076) & 0.944 & 1.101 \\    \hline \\
	\end{tabular}}
	\caption{Average MAE in Scenario 1. For each pair $(T, N)$, the best result is shown in bold. An asterisk indicates that a given method is significantly better than the rest at level $\alpha=0.01$.}
	\label{tablemae1}
\end{table}

The results in Table \ref{tablemae1} are coherent with the ones in Table \ref{tableari1}, with the proposed method outperforming the remaining approaches in most of the settings. Specifically, Table \ref{tablemae1} indicates that the predictive accuracy of local models is as good as that of global models for $T=400$, but significantly worse for shorter lengths. Note that CPAGM obtained substantially better results than fitting one global model to all the series in the collection ($K=1$) and GMAP, which was expected, since these approaches do not take into account the existence of different underlying generating processes. 

Average results for Scenario 2 concerning MAE are displayed in Table \ref{tablemae2}. The proposed approach showed a similar behaviour than in Scenario 1 in terms of predictive accuracy, but, as with clustering effectiveness (see Table \ref{tableari2}), the differences with respect to the remaining techniques were more pronounced in Scenario 2.

\begin{table}
	\centering
	\resizebox{8cm}{!}{\begin{tabular}{ccccc}  \hline 
			$(T,N)$ & LM & CPAGM ($K=1$)    & GMFBC & GMAP  \\   \hline 
			$(50, 5)$ & 1.854 & \textbf{1.375}$^*$ (1.871) & 1.657 & 1.902 \\
			$(50, 10)$ & 1.855 & \textbf{1.333}$^*$  (1.885) & 1.616 & 1.888 \\   
			$(50, 20)$ & 1.856 & \textbf{1.183}$^*$ (1.905) & 1.625 & 1.838 \\   
			$(50, 50)$ & 1.857 & \textbf{1.153}$^*$  (1.901) & 1.647 & 1.898 \\   \hline 
			$(100, 5)$ & 1.670 & \textbf{1.185}$^*$  (1.871) & 1.492 & 1.756 \\  
			$(100, 10)$ & 1.665 & \textbf{1.173}$^*$  (1.891) & 1.553 & 1.667 \\   
			$(100, 20)$ & 1.683 & \textbf{1.148}$^*$  (1.898) & 1.578 & 1.890 \\   
			$(100, 50)$ & 1.683 & \textbf{1.147}$^*$ (1.903) & 1.590 & 1.884 \\  \hline 
			$(200, 5)$ & 1.615 & \textbf{1.191}$^*$ (1.884) & 1.507 & 1.613 \\   
			$(200, 10)$ & 1.628 &\textbf{1.168}$^*$ (1.899) & 1.558 & 1.772 \\  
			$(200, 20)$ & 1.635 & \textbf{1.156}$^*$ (1.902) & 1.591 & 1.852 \\  
			$(200, 50)$ & 1.631 & \textbf{1.152}$^*$ (1.906) & 1.624 & 1.866 \\   \hline
			$(400, 5)$ &   1.566 & \textbf{1.197}$^*$ (1.906) & 1.483 & 1.743 \\  
			$(400, 10)$ & 1.574 & \textbf{1.177}$^*$ (1.898) & 1.526 & 1.729 \\  
			$(400, 20)$ & 1.561 & \textbf{1.177}$^*$ (1.900) & 1.573 & 1.885 \\   
			$(400, 50)$ & 1.563 & \textbf{1.181}$^*$ (1.904) & 1.596 & 1.916 \\   \hline 
			$(1000, 5)$ & 1.486 & \textbf{1.219}$^*$ (1.885) & 1.394 & 1.898 \\   
			$(1000, 10)$ & 1.513 & \textbf{1.231}$^*$ (1.899) & 1.473 & 1.887 \\ 
			$(1000, 20)$ & 1.516 & \textbf{1.210}$^*$  (1.908) & 1.497 & 1.892 \\  
			$(1000, 50)$ & 1.505 & \textbf{1.205}$^*$ (1.902) & 1.516 & 1.881 \\    \hline \\
	\end{tabular}}
	\caption{Average MAE in Scenario 2. For each pair $(T, N)$, the best result is shown in bold. An asterisk indicates that a given method is significantly better than the rest at level $\alpha=0.01$.}
	\label{tablemae2}
\end{table}

Results in terms predictive accuracy for the noisy scenario of Section \ref{subsubsectionnoisyscenario} are given in Table \ref{tablemaenoisy2}. The corresponding scores indicate that method CPAGM significantly outperforms the remaining approaches in most cases, but the differences in terms of MAE error are less pronounced than in the original Scenario 2. In this regard, it is worth remarking that the corresponding noise generation mechanism (i.e., the way the coefficients of the corresponding uniform distribution are chosen) naturally produces a decrease in the average prediction error. In addition, note that the predictive effectiveness of the local approach (LM) is not affected by the uncertainty in the model coefficients, since this method fits a different model to each one of the time series in the set.

	\begin{table}
	\centering
	\resizebox{8cm}{!}{\begin{tabular}{ccccc}  \hline 
			$(T,N)$ & LM & CPAGM ($K=1$)    & GMFBC & GMAP  \\   \hline 
			$(50, 5)$ & 1.381  & \textbf{1.189}$^*$  (1.385) & 1.347 & 1.363 \\
			$(50, 10)$ & 1.383 & \textbf{1.188}$^*$  (1.381) & 1.343 & 1.359 \\   
			$(50, 20)$ & 1.381 & \textbf{1.120}$^*$ (1.381) & 1.339 & 1.368 \\   
			$(50, 50)$ & 1.385 & \textbf{1.096}$^*$  (1.389) & 1.352 & 1.383 \\   \hline 
			$(100, 5)$ & 1.336  & \textbf{1.115}$^*$  (1.366) & 1.281 & 1.344 \\  
			$(100, 10)$ & 1.334 & \textbf{1.093}$^*$  (1.384) & 1.307 & 1.335 \\   
			$(100, 20)$ & 1.333 & \textbf{1.089}$^*$  (1.385) & 1.315 & 1.373 \\   
			$(100, 50)$ & 1.330 & \textbf{1.083}$^*$  (1.382) & 1.328 & 1.385 \\  \hline 
			$(200, 5)$ & 1.292  & \textbf{1.089}$^*$ (1.367) & 1.263 & 1.412 \\   
			$(200, 10)$ & 1.300 & \textbf{1.092}$^*$  (1.380) & 1.289 & 1.487 \\  
			$(200, 20)$ & 1.288 & \textbf{1.090}$^*$  (1.377) & 1.294 & 1.545 \\  
			$(200, 50)$ & 1.297 & \textbf{1.086}$^*$ (1.383) & 1.308 & 1.545 \\   \hline
			$(400, 5)$  & 1.255 & \textbf{1.099}$^*$ (1.380) & 1.237 & 1.468 \\  
			$(400, 10)$ & 1.261 & \textbf{1.100}$^*$  (1.392) & 1.279 & 1.483 \\  
			$(400, 20)$ & 1.250 & \textbf{1.098}$^*$  (1.384) & 1.283 & 1.531 \\   
			$(400, 50)$ & 1.255 & \textbf{1.096}$^*$  (1.388) & 1.295 & 1.514 \\   \hline 
			$(1000, 5)$ & 1.210  & \textbf{1.105}$^*$  (1.396) & 1.250 & 1.448 \\   
			$(1000, 10)$ & 1.196 & \textbf{1.097}  (1.377) & 1.250 & 1.497 \\ 
			$(1000, 20)$ & 1.201 & \textbf{1.106}  (1.388) & 1.268 & 1.511 \\  
			$(1000, 50)$ & 1.202 & \textbf{1.101}  (1.389) & 1.278 & 1.521 \\    \hline \\
	\end{tabular}}
	\caption{Average MAE in Scenario 2 with noisy coefficients. For each pair $(T, N)$, the best result is shown in bold. An asterisk indicates that a given method is significantly better than the rest at level $\alpha=0.01$.}
	\label{tablemaenoisy2}
\end{table}

Table \ref{tableoutofsamplemae} contains the results for the analysis of Section \ref{eofsea1} (out-of-sample error in Scenario 2) in terms of prediction effectiveness (MAE). CPAGM exhibits a worse behaviour for the shortest values of $T$ in comparison with original Scenario 2, but the differences are minor, thus allowing the proposed approach to maintain a substantial advantage over its competitors.

\begin{table}
	\centering
	\resizebox{3.5cm}{!}{\begin{tabular}{cc}  \hline 		
			$(T,N)$  & Average MAE \\   \hline 
			$(100, 5)$   & 1.237  \\   
			$(100, 10)$   & 1.200  \\  
			$(100, 20)$   & 1.190  \\ 
			$(100, 50)$   & 1.165  \\  \hline
			$(200, 5)$   & 1.208 \\ 
			$(200, 10)$   & 1.185  \\  
			$(200, 20)$   & 1.180  \\  
			$(200, 50)$   & 1.173  \\  \hline
			$(400, 5)$   & 1.210  \\ 
			$(400, 10)$   & 1.197  \\ 
			$(400, 20)$  & 1.189  \\ 
			$(400, 50)$  & 1.176  \\   \hline
			$(1000, 5)$ & 1.207  \\  
			$(1000, 10)$  & 1.198  \\  
			$(1000, 20)$  & 1.186 \\  
			$(1000, 50)$  & 1.184  \\    \hline \\
	\end{tabular}}
	\caption{Average MAE in Scenario 2 for method CPAGM. Out-of-sample error was used to assign the series to the clusters.}
	\label{tableoutofsamplemae}
\end{table}

Average results for Scenario 3 in Section \ref{subsubsectionnl} (nonlinear global models) in terms of MAE are given in Table \ref{tablemae3}. According to these values, one could state that CPAGM exhibits a significantly lower forecasting error than the remaining approaches in many cases. In fact, the accuracy of the proposed method improves when increasing the number of series per process, which is reasonable, since more series per cluster result in a better approximation of the underlying global models. 

	\begin{table}
	\centering
	\resizebox{8cm}{!}{\begin{tabular}{ccccc}  \hline 
			$(T,N)$ & LM & CPAGM ($K=1$)    & GMFBC & GMAP  \\   \hline 
			$(50, 5)$ & \textbf{1.486} & 1.488 (1.500) & 1.492 & 1.505 \\
			$(50, 10)$ & 1.496 & \textbf{1.405}$^*$  (1.493) & 1.539 & 1.493 \\   
			$(50, 20)$ & 1.491 & \textbf{1.379}$^*$ (1.465) & 1.460 & 1.477 \\ \hline 
			$(100, 5)$ & 1.421 & \textbf{1.338}  (1.460) & 1.393 & 1.459 \\  
			$(100, 10)$ & 1.413 & \textbf{1.296}$^*$  (1.434) & 1.399 & 1.443 \\   
			$(100, 20)$ & 1.440 & \textbf{1.277}$^*$  (1.431) & 1.441 & 1.458 \\ \hline 
			$(200, 5)$ & 1.387 & \textbf{1.304} (1.454) & 1.334 & 1.448 \\   
			$(200, 10)$ & 1.380 &\textbf{1.291}$^*$ (1.431) & 1.399 & 1.441 \\  
			$(200, 20)$ & 1.381 & \textbf{1.246}$^*$ (1.385) & 1.448 & 1.429 \\ \hline
			$(400, 5)$ &   1.308 & \textbf{1.272} (1.448) & 1.338 & 1.434 \\  
			$(400, 10)$ & 1.303 & \textbf{1.127}$^*$ (1.388) & 1.374& 1.416 \\  
			$(400, 20)$ & 1.351 & \textbf{1.141}$^*$ (1.372) & 1.326 & 1.428 \\ \hline 
			$(1000, 5)$ & 1.263 & \textbf{1.161}$^*$ (1.376) & 1.383 & 1.392 \\   
			$(1000, 10)$ & 1.236 & \textbf{1.125}$^*$ (1.367) & 1.331 & 1.385 \\ 
			$(1000, 20)$ & 1.245 & \textbf{1.083}$^*$  (1.362) & 1.375 & 1.402 \\ \hline 
	\end{tabular}}
	\caption{Average MAE in Scenario 3. For each pair $(T, N)$, the best result is shown in bold. An asterisk indicates that a given method is significantly better than the rest at level $\alpha=0.01$.}
	\label{tablemae3}
\end{table}

\subsection*{Combining series from two datasets}

In order to analyze the performance of the CPAGM and its competitors in heterogeneous datasets, we created two new databases as indicated below.

\begin{itemize}
	\item \textbf{Combined Dataset 1}. This database combines the series of the collections M1 Yearly and M3 Other, for a total of 355 time series. 
	\item \textbf{Combined Dataset 2}. This database combines the series of the collections M1 Quarterly and Tourism Quarterly, for a total of 630 time series. 
\end{itemize}

\begin{figure}
	\centering
	\includegraphics[width=0.7\textwidth]{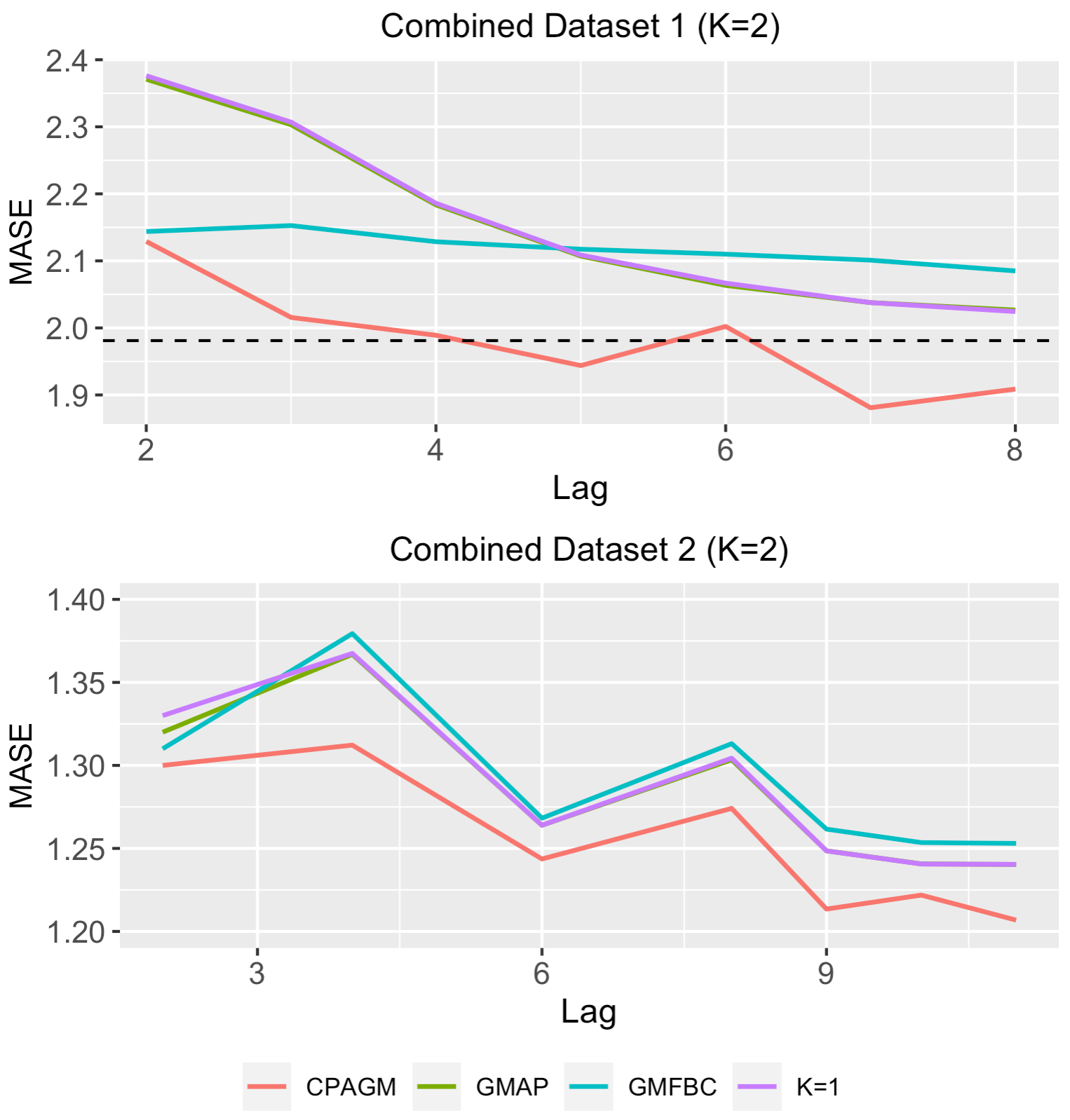}
	\caption{Average MASE as a function of the number of lags in datasets Combined Dataset 1 (top panel) and Combined Dataset 2 (bottom panel). A number of $K=2$ clusters was considered. Each color corresponds to a different method. The horizontal dashed line indicates the average MASE achieved by the approach based on local models (LM).}
	\label{plotcombined}
\end{figure}

The proposed algorithm and the alternative approaches were applied to perform clustering in the new datasets. This time, due to the nature of both databases, we set the number of clusters to $K=2$. The corresponding results in terms of average MASE are given in the top and bottom panels of Figure \ref{plotcombined} for Combined Dataset 1 and Combined Dataset 2, respectively. For each approach, a curve representing the MASE error as a function of the number of lags was displayed. As in Section \ref{subsectionapplication1}, the horizontal dashed line in the top panel indicates the average forecasting error associated with the LM approach. In Combined Dataset 1, the proposed method clearly outperforms the alternative clustering techniques and also the local approach when a sufficiently large order is considered for the global models. In Combined Dataset 2, CPAGM exhibits the lowest forecasting error over the whole range of lags. Note that, in both cases, the feature-based technique GMFBC shows a greater prediction error than the approach based on a single global model ($K=1$). This means that, in this challenging settings, the clusters detected by GMFBC are not able to properly describe the underlying forecasting structures. This highlights the importance of splitting the dataset in a way which is specifically designed to minimize the forecasting error instead of in a purely empirical manner. 

A numerical summary of Figure \ref{plotcombined} is given in the upper part of Table \ref{tablecomnbined}, where the average MASE associated with the optimal pair $(K,l)$ with $K=2$ is provided for each one of the approaches CPAGM, GMFBC and GMAP. The average forecasting error for the local approach LM was also included. The corresponding quantities corroborate the superiority of CPAGM over the remaining approaches, specially in Dataset Combined 1. Results in terms of sMAPE are provided in the middle part of Table \ref{tablecomnbined}. According to this error metric, the approach LM exhibits the best performance in the first dataset, but CPAGM still achieves the lowest forecasting error in the second database. Note that, in both cases, the differences with respect to the alternative clustering techniques (GMFBC) and (GMAP) are substantial.  

\begin{table}
	\centering
	\resizebox{11cm}{!}{\begin{tabular}{cccccc}\hline 
	Measure	& Dataset    & LM & CPAGM ($K=1$) & GMFBC & GMAP \\ \hline 
	MASE	& Combined Dataset 1 &  1.981   & \textbf{1.894} (2.024)  &  2.085     &  2.023     \\
		& Combined Dataset 2 &   2.096  &      \textbf{1.206} (1.240)      &  1.253     &  1.240      \\ \hline 
	sMAPE	& Combined Dataset 1 &  \textbf{24.419}   &   40.675 (67.570) &   67.371    &  63.942    \\
		& Combined Dataset 2 &  32.543   &    \textbf{27.307} (33.535)         &    40.216   &    40.860   \\ \hline 
	ARI	& Combined Dataset 1 &  0.028   &  0.312 (-) &    \textbf{0.548}   &   -   \\
	& Combined Dataset 2  &         -0.014  &    \textbf{0.497}    (-)    &   0.096    &  -     \\ \hline 
	\end{tabular}}
	\caption{Performance of the different methods in terms of MASE (upper part), sMAPE (middle part) and ARI (lower part) in datasets Combined Dataset 1 and Combined Dataset 2. A number of $K=2$ clusters was considered. For each dataset and performance measure, the best result is shown in bold. The optimal value for $l$ was considered for each method and dataset.}
	\label{tablecomnbined}
\end{table}

To get insights into the clustering solutions produced by LM, CPAGM and GMFBC, we decided to compare the corresponding partitions with the ground truth, which was assumed to be defined by the two datasets constituting each one of the combined databases. As in the simulations of Section \ref{sectionsimulationstudy}, the estimated coefficients of the local models were used as input to the traditional $K$-means method $(K=2)$ in order to obtain the clustering solution associated with LM. The considered clustering solutions for method CPAGM were the ones related to the optimal values for $l$, namely $l=7$ in Combined Dataset 1 and $l=11$ in Combined Dataset 2 (see Figure \ref{plotcombined}). 

Values of ARI index are given in the lower part of Table \ref{tablecomnbined}. The clustering solutions produced by LM are not consistent with the true partitions, since the corresponding ARI values are close to zero. On the contrary, CPAGM displays moderately high values for the ARI in both cases, specially for Combined Dataset 2, which indicates that series of datasets M1 Quarterly and Tourism Quarterly get partially located in different groups of the experimental partition. The feature-based approach GMFBC also attains a rather high ARI value in Combined Dataset 1. In fact, this value is substantially higher than the ARI obtained by CPAGM in this database. However, as shown in the top panel of Figure \ref{plotcombined}, a higher ARI does not result in a better predictive accuracy. This is not surprising, since, unlike in the simulations of Section \ref{sectionsimulationstudy}, the definition of the true partition in this setting was done in a purely empirical way, and values of the ARI index are shown here only for exploratory purposes. In fact, according to the results for CPAGM, one can conclude that the optimal partition in terms of forecasting error is not defined by both individual datasets, although a certain degree of similarity between both partitions exists. This is reasonable, since two time series from the same domain, although somehow related, do not necessarily share the same forecasting structure. 

In short, the analyses carried out throughout Sections \ref{subsectionapplication1} and \ref{subsectionapplication2} show the usefulness of the proposed method when dealing with real time series datasets. Particularly, CPAGM generally outperforms alternative approaches in terms of prediction error, which makes this method an appropriate choice to carry out forecasting tasks in time series databases.

	% that's all folks
\end{document}